%% file: eccv2022submission.tex
\documentclass[runningheads]{llncs}
\usepackage{graphicx}

\usepackage{tikz}
\usepackage{comment}
\usepackage{amsmath,amssymb} 
\usepackage{color}

\usepackage[accsupp]{axessibility}  

\usepackage{xspace}
\usepackage{graphicx}

\usepackage{booktabs}

\usepackage{float}
\usepackage{todonotes}

\usepackage[marginal]{footmisc}

\usepackage[pagebackref,breaklinks,colorlinks]{hyperref}
\usepackage{algorithm, algorithmic}
\usepackage{cite}
\usepackage{caption, subcaption}
\usepackage[capitalize]{cleveref}

\crefname{section}{Sec.}{Secs.}
\Crefname{section}{Section}{Sections}
\Crefname{table}{Table}{Tables}
\crefname{table}{Tab.}{Tabs.}

\newcommand{\sys}{MegBA\xspace}

\newcommand*{\myc}[1]{%
\scalebox{0.78}{\begin{tikzpicture}[baseline=-3pt]
  \protect\node[draw,circle,inner sep=0.5pt,fill=black] at (0, 0) {\textcolor{white}{\textsf{\textbf{#1}}}};
\end{tikzpicture}}}

\begin{document}
\newcommand\thelinenumber{\color[rgb]{0.2,0.5,0.8}\normalfont\sffamily\scriptsize\arabic{linenumber}\color[rgb]{0,0,0}}
\newcommand\makeLineNumber {\hss\thelinenumber\ \hspace{6mm} \rlap{\hskip\textwidth\ \hspace{6.5mm}\thelinenumber}}

\newcommand{\Comment}[1]{%
  \leavevmode\hfill{\textit{/*~#1~*/}}}
\pagestyle{headings}
\mainmatter
\def\ECCVSubNumber{4366}  

\title{\sys: A GPU-Based Distributed Library for Large-Scale Bundle Adjustment} 

\titlerunning{\sys}
%
\author{Jie Ren$^{*}$\inst{1,2}\orcidID{0000-0003-3332-2092} \and
Wenteng Liang$^{*}$\inst{1}\orcidID{0000-0002-3750-7695} \and
\\ Ran Yan$^{\dag}$\inst{1}\orcidID{0000-0001-6705-1644} \and
Luo Mai\inst{2}\orcidID{0000-0002-3594-1092} \and
\\ Shiwen Liu\inst{1}\orcidID{0000-0003-1779-409X} \and
Xiao Liu\inst{1}\orcidID{0000-0002-6073-030X}}
\authorrunning{J. Ren, et al.}

%
\institute{Megvii Inc., China \and
University of Edinburgh, United Kingdom}
\maketitle
\footnote{$^{*}$ Equal contribution, work was done during their internship in Megvii Inc.\\
$^{\dag}$ Corresponding author.}
\begin{abstract}
Large-scale Bundle Adjustment (BA) requires massive memory and computation resources which are difficult to be fulfilled by existing BA libraries. In this paper, we propose \sys, a GPU-based distributed BA library. \sys can provide massive aggregated memory by automatically partitioning large BA problems, and assigning the solvers of sub-problems to parallel nodes. The parallel solvers adopt distributed Precondition Conjugate Gradient and distributed Schur Elimination, so that an effective solution, which can match the precision  of those computed by a single node, can be efficiently computed. To accelerate BA computation, we implement end-to-end BA computation using high-performance primitives available on commodity GPUs. \sys exposes easy-to-use APIs that are compatible with existing popular BA libraries. Experiments show that MegBA can significantly outperform state-of-the-art BA libraries: Ceres (41.45$\times$), RootBA (64.576$\times$) and DeepLM (6.769$\times$) in several large-scale BA benchmarks. The code of \sys is available at: \url{https://github.com/MegviiRobot/MegBA}.

\end{abstract}

\input{src/1intro}
\input{src/2RelatedWork}
\input{src/3preliminary}

\input{src/4method}
\input{src/5GPU}
\input{src/6Experiment}

\section{Conclusion}
We present \sys, a novel GPU-based distributed BA library. \sys has a set of algorithms that enables automatically distributing BA computation to parallel GPUs. It has a group of SIMD-optimised data structures, and a memory-efficient runtime, making \sys capable of fully utilising a GPU. \sys has high-level and compatible APIs, making it quickly become a popular open-sourced BA library. Experimental results show that \sys can out-perform SOTA BA libraries by orders of magnitudes in several large-scale BA benchmarks.

\section*{Acknowledgement}
We would like to thank our colleagues at Megvii Inc., Dikai Fan, Shuxue Peng, Yijia He, Zheng Chai, Haotian Zhang and Can Huang for their invaluable inputs. Also, we thank Qingtian Zhu at Peking University and Xingxing Zuo at Technical University of Munich to help with proof reading.

\clearpage
%
%

\bibliographystyle{splncs04}
\bibliography{egbib}

\clearpage
\input{supp}

\end{document}

%% file: src/1intro.tex
\section{Introduction}
Bundle Adjustment (BA) is the foundation for many real-world 3D vision applications~\cite{BA-modern, FrameSLAM}, including structure-from-motion and simultaneous-localization-and-mapping. A BA problem minimises the re-projection error between camera poses and map points. The error is a non-linear square function, and it is minimised through iterative methods, such as Gauss-Newton (GN)~\cite{GN}, Leverberg-Marquardt (LM)~\cite{LM} and Dog-Leg~\cite{Dog-Leg}. In each iteration, a BA library differentiates the errors with respect to  solving states and constructs a linear system which is solved by optimisation algorithms, such as Cholesky decomposition~\cite{CHOLMOD} and Precondition Conjugate Gradient (PCG)~\cite{PCG}.

Large-scale BA is increasingly important given the recent rise of city-level high-definition maps for autonomous driving~\cite{AD1, AD2, AD3, AD4} and indoor maps for augmentation reality~\cite{AR1, AR2, AR3}. A structure-from-motion application, for example, can produce massive images~\cite{dataset1, BAL}, resulting in billions of points and observations to be adjusted. Such a BA problem is orders of magnitude larger than those in conventional vision applications~\cite{stba, reconstruction2}. 


Existing BA libraries (e.g., g2o~\cite{g2o} and Ceres~\cite{ceres}) however provide insufficient support for large-scale BA. We observe several reasons: \textbf{(i)}~Existing libraries focus on single-node execution, and they lack algorithms to distribute computation. They thus cannot provide massive aggregated memory that is the key for large-scale BA. Even though there are algorithms, such as RPBA~\cite{rpba}, DPBA~\cite{dpba} and STBA~\cite{stba}, which explore distributed BA. These algorithms adopt \emph{approximation} which can adversely affect the precision of found solutions. \textbf{(ii)}~Existing BA libraries are designed for CPU architectures, and they under-utilise GPUs which is particularly useful for large-scale BA. Even though there are systems, such as PBA~\cite{pba}, to accelerate BA with GPUs. They leave key BA operations un-accelerated (e.g., error differentiation and linear system construction). DeepLM~\cite{DeepLM} offloads error differentiation into GPUs through PyTorch, but the performance is often sub-optimised.

In this paper, we propose \sys, a novel GPU-optimised distributed library for large-scale BA. The design of \sys makes several contributions:

\noindent
\textbf{(1) Distributed BA algorithms}. \sys provides a large amount of aggregated memory by distributing BA computation to multiple nodes.
To this end, we propose a generic BA problem partitioning method. This method leverages a key observation in BA problems: BA problems are often expressed as graphs where nodes represent points/cameras, and edges represent the associations between cameras and points. \sys can thus automatically partition the graphs based on edges, and ensure each sub-graph has an equal number of edges (with an aim of achieving load balancing). \sys further assigns sub-graphs to distributed nodes and merges the local solutions to sub-graphs.
To ensure that distributed BA can offer the precision as those computed by single-node BA libraries, we propose the distributed PCG algorithm and the distributed Schur elimination algorithm. These two algorithms  synchronise the states of solvers on parallel nodes, and the synchronisation is realised using NCCL.

\noindent
\textbf{(2) GPU-Optimised BA computation}. 
\sys thoroughly optimise BA computation for GPUs, thus providing massive computation power for large-scale BA. Computation-intensive operators (e.g., inverse, inner project, etc) are implemented as Single-Instruction-Multiple-Data (SIMD) operators. \sys store data in \emph{JetVector}, a data structure that stores BA data in SIMD-friendly vectors, and JetVector minimises data serialisation cost between CPUs and GPUs. To minimise data movement cost which could block GPU execution, \sys has algorithms that can predict the GPU memory usage of BA, thus pre-fetching BA data if possible. It exposes easy-to-use APIs that are compatible with g2o and Ceres. Ceres and g2o applications can be thus easily ported to \sys. 

We evaluate the performance of \sys on servers and each server has 8 NVIDIA V100 GPUs. Experiments with public large BA datasets (i.e., Final-13682~\cite{BAL}) show that MegBA can out-perform Ceres by up to 41.45$\times$, and RootBA~\cite{RootBA} by up to 64.576$\times$, indicating the benefits of optimising BA computation for GPUs. We further compare MegBA with DeepLM~\cite{DeepLM}, a GPU-based BA library. MegBA out-performs DeepLM by 5.213$\times$ on 4 GPUs. With 8 GPUs, MegBA out-performs DeepLM by 6.769$\times$, making \sys the state-of-the-art BA library on GPUs. 

To evaluate the scalability of \sys, we construct an extremely large synthetic BA dataset which is modelled after by the BA problems we have in real-world applications.
This dataset contains 80 million observations, 2.76$\times$ larger than Final-13682. DeepLM and RootBA incur out-of-memory error and cannot handle such a dataset. On the contrary, \sys can solve this BA problem in 216.26 seconds by distributing BA computation to 8 GPUs, which is 23.54$\times$ faster than Ceres.

%% file: src/2RelatedWork.tex
\section{Related Work}

This section describes the related work of \sys. g2o~\cite{g2o} and Ceres~\cite{ceres} are \textbf{exact BA libraries} that can compute high-accuracy solutions to BA problems. These libraries are designed for using parallel CPU cores, and they cannot use GPUs. These libraries also fail to provide distributed execution, which makes them suffer from out-of-memory issues in solving large-scale BA problems.

\textbf{Approximated BA algorithms} can substantially speed up BA, though often come with a compromise in the quality of BA solutions. PBA~\cite{pba} is limited to run on a single device. $\sqrt{BA}$ \cite{demmel2021square} replaced Schur Complement with a memory-efficient nullspace projection of Jacobian, thus improving its performance with single-precision float numbers. iSAM\cite{iSAM} and iSAM2\cite{iSAM2} exploit states ordering; while ICE-BA \cite{ICE-BA} exploits the states in temporal orders. Though fast in speed, approximated BA algorithms modify the original BA problems, which adversely affect the quality of BA solutions. As a result, commercial 3D vision software, such as PIX4D$^1$ usually avoid any form of approximation and adopt exact BA libraries if possible. \footnote{$^1$https://www.pix4d.com/}

\textbf{Distributed BA libraries} have been recently designed for large-scale BA. Anders Eriksson et al.~\cite{consensus1} present consensus-based optimisation which leverages proximal splitting. Runze Zhang et al.~\cite{dpba_gcc} purpose an Alternating Direction Method of Multipliers to distribute the optimisation problem. 
Later RPBA~\cite{rpba}, DPBA~\cite{dpba}, STBA~\cite{stba} partition the BA problems based on ADMM. These ADMM-based systems incur massive redundant computation on distributed devices, making them sometimes under-perform single-node libraries. Further, their users must manually partition BA problems, resulting in sub-optimal distributed performance. BA-Net \cite{BA-Net} and DeepLM \cite{DeepLM} leverages GPUs to speed up BA. They however rely on PyTorch to use GPU, which incurs non-trivial performance overheads when using GPU and extra memory copies.
Decentralised SLAM libraries, such as DEDV-SLAM~\cite{DEDVSLAM}, often solve approximated BA problems on distributed robots, then they merge local solutions. However, the merged solution is not equivalent to the original global BA problem.

\textbf{Custom hardware and algorithms} are useful in accelerating BA~\cite{guo2021fast}. GBP~\cite{IPU} uses a neural processing unit (i.e., GraphCore IPU) to speed up BA; but the limited availability of IPU makes GBP difficult to be used as a general solution. Practitioners also propose an approximated BA solver tailored for facial capture~\cite{Disney}, and this solver cannot be used for arbitrary BA problems such as structure-from-motion. 

%% file: src/3preliminary.tex
\section{Preliminaries}
\input{src/pipeline}
This section introduces the preliminaries of \sys. 
A BA problem can be expressed as a graph, and its solving is realised an iterative process which minimises a non-linear square error objective function:
\begin{equation}
\boldsymbol{x}^* = \arg\min_{\boldsymbol{x}} \sum \boldsymbol{e}_{i, j}^\top \boldsymbol{\Sigma}_{i, j} \boldsymbol{e}_{i, j}, 
\label{eq:obj}
\end{equation}

\noindent
where $\boldsymbol{e}_{i, j}$ is the constraint (i.e. error or graph edge) between state (i.e. parameters or graph nodes) $\boldsymbol{x}_{i}$ and $\boldsymbol{x}_{j}$, $\boldsymbol{\Sigma}_{i, j}$ is an information matrix. 

Solving Equation~\ref{eq:obj} is equivalent to iteratively updating the incremental amount $\boldsymbol{\Delta x}$, given by the linear system $ \boldsymbol{H} \Delta \boldsymbol{x} = \boldsymbol{g} $, upon the current state $\boldsymbol{x}$ until convergence. 
The Hessian matrix $\boldsymbol{H} = \boldsymbol{J}^{T}\Sigma \boldsymbol{J}$ for GN method and $ \boldsymbol{H} = \boldsymbol{J}^{T}\Sigma \boldsymbol{J} + \lambda \boldsymbol{I}$ for LM method, the residual vector $\boldsymbol{g}$ equals to $-\boldsymbol{J}^{T}\Sigma \boldsymbol{r}$, $\boldsymbol{J}$ is the Jacobian of the error $\boldsymbol{e}$ with respect to current state $\boldsymbol{x}$. 

To solve BA problems, BA libraries can use Schur Complement (SC):
\begin{equation}
\begin{bmatrix}
\boldsymbol{B} & \boldsymbol{E} \\ 
\boldsymbol{E}^{T} & \boldsymbol{C} 
\end{bmatrix} 
\begin{bmatrix}
\Delta \boldsymbol{x}_{c}\\ 
\Delta \boldsymbol{x}_{p}
\end{bmatrix} = \begin{bmatrix}
\boldsymbol{v}\\ 
\boldsymbol{w}
\end{bmatrix}
\label{eq:schur_complement}
\end{equation}

\noindent
where $\boldsymbol{B}$ and $\boldsymbol{C}$ are block diagonal and they are related to camera-camera and point-point edges, respectively. $\boldsymbol{E}$ refers to edges between camera and point. $\boldsymbol{v}$ and $\boldsymbol{w}$ refer to the residual vectors for camera and point states.

Solving $\boldsymbol{H}\Delta \boldsymbol{x} = \boldsymbol{g}$ is equivalent to compute the incremental update for states related to cameras $\Delta \boldsymbol{x}_{c}$ by solving an alternative linear system, called Reduced Camera System (RCS) \begin{equation}
[\boldsymbol{B} - \boldsymbol{E}\boldsymbol{C}^{-1}\boldsymbol{E}^T]\Delta \boldsymbol{x}_c = \boldsymbol{v} - \boldsymbol{E}\boldsymbol{C}^{-1}\boldsymbol{w}, 
\label{eq:schur_x_c}
\end{equation}
and followed by a substitution $\Delta \boldsymbol{x}_{c}$ into \begin{equation}
\Delta \boldsymbol{x}_{p} = \boldsymbol{C}^{-1}\left (\boldsymbol{w} - \boldsymbol{E}^{T} \Delta \boldsymbol{x}_{c}  \right ),
\label{eq:schur_x_p}
\end{equation}
to get the update for 3D map points.

BA libraries solve linear systems using either direct methods or iterative methods. Direct methods, such as Gaussian-Elimination, LU, QR, and Cholesky Decomposition, return optimised solution of $\boldsymbol{x}$ in one pass. They however suffer from $O(n^3)$ time and $O(n^2)$ space complexity, making them only suitable for small-scale BA problems. On the contrary, iterative methods, such as PCG~\cite{BA-modern}, are suitable for large-scale BA problems. Specifically, PCG replaces the explicit computation of $\boldsymbol{E}\boldsymbol{C}^{-1}\boldsymbol{E}^T$ with multiple iterative sparse matrix-vector operations. It reduces the space complexity to $O(n)$, thus saving memory.

%% file: src/pipeline.tex
\begin{figure*}[t!]
\centering
\includegraphics[width=\textwidth]{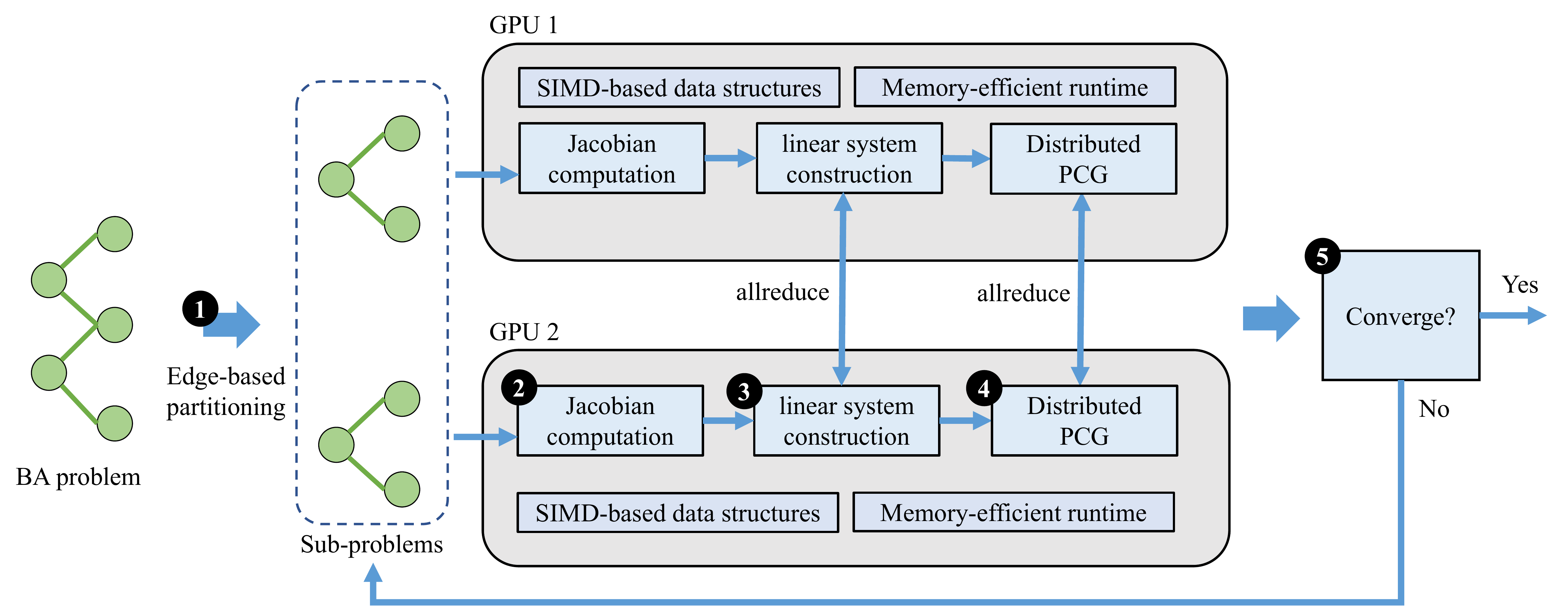}
\caption{\textbf{MegBA overview}. \textit{\myc{1} \sys partitions a BA problem based on edges. BA sub-problems are in the same size, and they are dispatched to distributed GPUs. Each GPU computes Jacobians \myc{2}, constructs a linear system \myc{3}, and solve the linear system using the the distributed PCG algorithm \myc{4}. The communication involved in linear system construction and distributed PCG is implemented using allreduce operations. Step \myc{2}, \myc{3}, and \myc{4} are executed iteratively until \myc{5} convergence criteria has been met.} }
\label{fig:pipeline}
\end{figure*}

%% file: src/4method.tex
\section{\sys Design}
This section introduces the design of \sys. A key design goal of \sys is to transparently distribute the solving of a given BA problem to multiple nodes, thus addressing the memory wall of a single node.

Figure~\ref{fig:pipeline} presents an overview of the distributed execution of \sys.  A \sys user declares a BA problem as a graph. 
\sys can automatically partition the BA problem based on edges with an aim of each BA sub-problem to have an even number of edges \myc{1}. Specifically, each GPU first \myc{2} computes the Jacobian (i.e. differentiation of the edge for the node), and then \myc{3} construct the linear system, and finally, \myc{4} apply PCG to compute the update for adjusting the current BA sub-problem. The PCG intermediate state is synchronised so that \sys can eventually solve the shared global BA problem. The BA update step is iteratively performed until a user-defined convergence criterion is met \myc{5}. 
Notably, the BA computation on each GPU is implemented as SIMD operations which can best utilise GPUs (The details of SIMD-friendly data structure and the memory-efficient runtime are given in Section~\ref{sec:implementation})

\subsection{Edge-Based Partitioning Method}

We focus on partitioning the Hessian matrix produced in BA. For example, in a BA dataset with 80M edges, a Hessian matrix $\boldsymbol{H}$ consume over 50G memory, leading to over 99.9\% storage to be allocated to $\boldsymbol{E}$ and $\boldsymbol{E}^T$. 
We want to have a generic method that partitions the Hessian matrix in a BA problem. This method needs to assign each parallel device with a part of the matrix $\boldsymbol{E}$ and $\boldsymbol{E}^T$, preferably in equal sizes. The partitioning needs to guarantee that the global solution merged from local solutions is \emph{equivalent} to the one computed using a single node. This equivalence property is the key to ensuring a high-precision solution found by \sys.

At the high level, \sys achieves distributed BA using two major components:
(i)~a method that can divide a BA problem into sub-problems, and (ii)~an algorithm that can coordinate distributed PCG algorithms to solve sub-problems in parallel. Our partitioning method is based on a key observation in BA problems: the non-zero blocks in $\boldsymbol{E}$ and $\boldsymbol{E}^T$ are corresponding with edges, i.e., the $i$-th row $j$-th column non-zero block in $\boldsymbol{E}$ is computed by $e_{i, j}$, we can partition edges based on the number of available GPUs, and each GPU only store part of these non-zero blocks (We provide an example to illustrate the partitioning process in the supplementary materials). 

Assume there are $K$ GPUs, given a BA problem, we tile edges to a vector $\boldsymbol{e} = \begin{array}{ccc}[\dots & e_{i, j} & \dots]^T\end{array}$, then we partition it to several blocks $\boldsymbol{e} = \begin{array}{cccc}[\boldsymbol{e}_1^T & \boldsymbol{e}_2^T & \dots & \boldsymbol{e}_{K}^T]^T\end{array}$. The Jacobian $\boldsymbol{J}$ could be partitioned into several blocks:

\begin{equation}
    \boldsymbol{J}
    = \frac{d\boldsymbol{e}}{d\boldsymbol{x}}
    = \left[
    \begin{array}{cccc}
    \frac{d\boldsymbol{e}_1}{d\boldsymbol{x}}^T & \frac{d\boldsymbol{e}_2}{d\boldsymbol{x}}^T & \dots & \frac{d\boldsymbol{e}_K}{d\boldsymbol{x}}^T
    \end{array} \right]^{T}
    = \left[
    \begin{array}{cccc}
    \boldsymbol{J}_1^T & \boldsymbol{J}_2^T & \dots & \boldsymbol{J}_{K}^T
    \end{array} \right]^{T}.
\end{equation}

Assuming identity information matrix is given here, Hessian $\boldsymbol{H}$ can be partitioned:
\begin{equation}
    \begin{aligned}
    \boldsymbol{H} &= \boldsymbol{J}^T\boldsymbol{J} &= \sum_{k=1}^K \boldsymbol{J}_k^T\boldsymbol{J}_k = \sum_{k=1}^K \boldsymbol{H}_k.
    \end{aligned}
\end{equation}

To perform Schur elimination, we represented $\boldsymbol{H}_k$ as sub-blocks:
\begin{equation}
    \boldsymbol{H}_k = \left[
    \begin{array}{cc}
    \boldsymbol{B}_k     &  \boldsymbol{E}_k \\
    \boldsymbol{E}_k^T   & \boldsymbol{C}_k
    \end{array}
    \right].
\end{equation}

Matrix blocks in Equation~\ref{eq:schur_complement} have the following equivalent forms in the edge-based partition setting:  $\boldsymbol{B} = \sum_{k=1}^K \boldsymbol{B}_k$, $\boldsymbol{E} = \sum_{k=1}^K \boldsymbol{E}_k$, $\boldsymbol{E}^T = \sum_{k=1}^K \boldsymbol{E}_k^T$, and $\boldsymbol{C} =\sum_{k=1}^K \boldsymbol{C}_k$. The number of non-zero parameter blocks in $\boldsymbol{E}$ or $\boldsymbol{E}^T$ equals the number of edges. Notably, the sub-matrices $\boldsymbol{B}_k$ and $\boldsymbol{C}_k$ have the same number of non-zero elements as $\boldsymbol{B}$ and $\boldsymbol{C}$, respectively. Since we store matrices in the Compressed Sparse Row (CSR) format, each GPU only stores $\frac{1}{K}$ non-zero blocks in $\boldsymbol{E}$ and $\boldsymbol{E}^T$. The blocking strategy greatly alleviates the problem that $\boldsymbol{E}$ and $\boldsymbol{E}^T$ are too large to be stored on a single device.

By applying the above partition method for Equation~\ref{eq:schur_complement}, an equivalent distributed version can be formulated as follow:
\begin{equation}
    \boldsymbol{g} = -\boldsymbol{J}^T\boldsymbol{r}
    = -[\begin{array}{cccc}
        \boldsymbol{J}_1^T & \boldsymbol{J}_2^T & \dots \boldsymbol{J}_K^T
    \end{array}]
    [\begin{array}{cccc}
        \boldsymbol{r}_1^T & \boldsymbol{r}_2^T & \dots & \boldsymbol{r}_K^T
    \end{array}]^T
    = -\sum_{k=1}^K\boldsymbol{J}_k^T\boldsymbol{r}.
\end{equation}
\label{sec:distributed_method}

\subsection{Distributed BA Algorithm}


\begin{algorithm}[t]
\caption{\textbf{Distributed BA}}
\label{alg:distributed_BA}
\begin{algorithmic}[1]
\REQUIRE BA initial state $\boldsymbol{x} = [\begin{array}{cc}
    \boldsymbol{x}_c^T & \boldsymbol{x}_p^T 
\end{array}]^T$, vector of edges $\boldsymbol{e}_k$, and local GPU rank $k$
\ENSURE Optimised state $\boldsymbol{x}$
\WHILE{\textit{BA Convergence Criteria} not satisfied}
    \STATE $\boldsymbol{r}_k = \boldsymbol{e}_k(\boldsymbol{x}), \boldsymbol{J}_k = d\boldsymbol{e}_k(\boldsymbol{x})/d\boldsymbol{x}$ \Comment{Residual and Jacobian}\label{alg:distributed_BA:JV}

    \STATE $\left[\begin{array}{cc}
        \boldsymbol{B}_k & \boldsymbol{E}_k \\
        \boldsymbol{E}_k^T & \boldsymbol{C}_k
    \end{array}\right] = \boldsymbol{J}_k^T\boldsymbol{J}_k, [\begin{array}{cc}
        \boldsymbol{v}_k & \boldsymbol{w}_k
    \end{array}] =  -\boldsymbol{J}_k^T\boldsymbol{r}_k$\Comment{Hessian and Constant vector}\label{alg:distributed_BA:build_linear}

    \STATE $\boldsymbol{B} = allreduce(\boldsymbol{B}_k)$, 
    $\boldsymbol{C} = allreduce(\boldsymbol{C}_k)$,\\
    $\boldsymbol{v} = allreduce(\boldsymbol{v}_k)$, 
    $\boldsymbol{w} = allreduce(\boldsymbol{w}_k)$\\
    \Comment{
    $\boldsymbol{B} = \sum_{i=1}^K\boldsymbol{B}_i$,
    $\boldsymbol{C} = \sum_{i=1}^K\boldsymbol{C}_i$,
    $\boldsymbol{v} = \sum_{i=1}^K\boldsymbol{v}_i$,
    $\boldsymbol{w} = \sum_{i=1}^K\boldsymbol{w}_i$
    }\label{alg:distributed_BA:reduce_linear}
    \STATE $\boldsymbol{\alpha}_k = \boldsymbol{E}_k\boldsymbol{C}^{-1}\boldsymbol{w}$
    \label{alg:distributed_BA:reduce_constant_start}
    \STATE $\boldsymbol{\alpha} = allreduce(\boldsymbol{\alpha}_k)$
    \Comment{$\sum_{i=1}^K\boldsymbol{\alpha}_i$}
    \STATE $\boldsymbol{g} = \boldsymbol{v} - \boldsymbol{\alpha}$\Comment{Constant vector in Equation~\ref{eq:schur_x_c}}\label{alg:distributed_BA:reduce_constant_end}
    \STATE$\Delta\boldsymbol{x}_c = \texttt{DPCG}(\boldsymbol{0}, \boldsymbol{B}, \boldsymbol{E}_k, \boldsymbol{E}_k^T, \boldsymbol{C}, \boldsymbol{g}, k)$\Comment{Update $\boldsymbol{x}_c$ using Algorithm~\ref{alg:DPCG}}\label{alg:distributed_BA:solve_schur}

    \STATE $\boldsymbol{\beta}_k = \boldsymbol{E}_k^T\Delta\boldsymbol{x}_c$
    \label{alg:distributed_BA:compute_xp_start}
    \STATE $\boldsymbol{\beta} = allreduce(\boldsymbol{\beta}_k)$
    \Comment{$\sum_{i=1}^K\boldsymbol{\beta}_i$}
    \STATE$\Delta \boldsymbol{x}_{p} = \boldsymbol{C}^{-1}(\boldsymbol{w} - \boldsymbol{\beta})$
    \Comment{Increment of $\boldsymbol{x}_p$}
    \label{alg:distributed_BA:compute_xp_end}
    \STATE $\boldsymbol{x}_c = \boldsymbol{x}_c + \Delta\boldsymbol{x}_c$, $\boldsymbol{x}_p = \boldsymbol{x}_p + \Delta\boldsymbol{x}_p$\Comment{Update state}
    \label{alg:distributed_BA:update}
\ENDWHILE
\RETURN $\boldsymbol{x} = [\begin{array}{cc}
    \boldsymbol{x}_c^T & \boldsymbol{x}_p^T
\end{array}]^T$
\end{algorithmic}
\end{algorithm}

By far we have partitioned a BA problem and assigned sub-problems to all GPUs. In the following, we will discuss how does \sys coordinates the solving of sub-problems in a distributed manner. 

Algorithm~\ref{alg:distributed_BA} introduces the overall distributed BA algorithm in \sys.  The distributed BA algorithm takes as initial state and partitioned edges as described in Section~\ref{sec:distributed_method}. We use \emph{JecVector} to compute the Jacobian and residual (Line~\ref{alg:distributed_BA:JV}). \emph{JetVector} is a novel data structure to represent BA data in a SIMD format, it can make full use of the hardware characteristics of GPU (e.g. coalesced memory loading) to do auto-differentiation over millions of edges in parallel. We give a more detailed illustration in Section~\ref{sec:JV}. Then we build a linear system (Line~\ref{alg:distributed_BA:build_linear}). We perform allreduce~\cite{mai2020kungfu} on diagonal-blocks and constant vector (Line~\ref{alg:distributed_BA:reduce_linear}) before solving the linear system because the size of diagonal-blocks and constant vector is small and they would be used several times in the following procedures. 

We then compute constant vector in Equation~\ref{eq:schur_x_c} (Line~\ref{alg:distributed_BA:reduce_constant_start}-\ref{alg:distributed_BA:reduce_constant_end}) and solve the linear system by using a Distributed PCG (DPCG) algorithm (Line~\ref{alg:distributed_BA:solve_schur}). Notably, we do necessary allreduce in the DPCG algorithm to guarantee DPCG output the same result as non-distributed PCG solver does in solving Equation~\ref{eq:schur_x_c}, further implementation details will be shown in Section~\ref{sec:DPCG}. After solving the linear system in Equation~\ref{eq:schur_x_c}, we compute the increment of $\boldsymbol{x}_p$ following Equation~\ref{eq:schur_x_p} (Line~\ref{alg:distributed_BA:compute_xp_start}-\ref{alg:distributed_BA:compute_xp_end}). Once we have computed the increment $\Delta\boldsymbol{x}_c$ and $\Delta\boldsymbol{x}_p$, we update the state $\boldsymbol{x}_c$ and $\boldsymbol{x}_p$ (Line~\ref{alg:distributed_BA:update}). If it doesn't satisfy the convergence criteria we will start another loop; otherwise, we will return the optimised state $\boldsymbol{x}$.

\begin{algorithm}[t!]
\caption{\textbf{Distributed PCG (DPCG)}}
\label{alg:DPCG}
\begin{algorithmic}[1]
\REQUIRE Initial state $\boldsymbol{x}^{0}$, matrix block $\boldsymbol{B}$, $\boldsymbol{E}_k$, $\boldsymbol{E}_k^T$, $\boldsymbol{C}$ of $\boldsymbol{H}_k$, constant vector $\boldsymbol{b}$, and local GPU rank $k$
\ENSURE Solution $\boldsymbol{x}$ for linear system [$\boldsymbol{B} - \boldsymbol{E}\boldsymbol{C}^{-1}\boldsymbol{E}^T] \boldsymbol{x} = \boldsymbol{b}$, where $\boldsymbol{E} = \sum_{i=1}^K\boldsymbol{E}_i$ and $\boldsymbol{E}^T = \sum_{i=1}^K\boldsymbol{E}_i$

\STATE $ \boldsymbol{r}^{0} = \boldsymbol{b} - \texttt{DSE}(\boldsymbol{x}^0, \boldsymbol{B},  \boldsymbol{E}_k, \boldsymbol{E}_k^T, \boldsymbol{C}^{-1}, k)$ \Comment{Algorithm~\ref{alg:DSE}} \label{alg:DPCG:MV1}

\STATE $ n = 0 $
\WHILE{\textit{Convergence Criteria} not satisfied}
    \STATE $ \boldsymbol{z}^{n} = \boldsymbol{B}^{-1}\boldsymbol{r}^{n}$   
    \STATE $\rho^{n} = {\boldsymbol{r}^{n}}^T\boldsymbol{z}^{n} $  \label{DPCG:dot1} 
    \IF{$ n > 1 $}
        \STATE $ \beta^{n} = \rho^{n} / \rho^{n-1}$
        \STATE $ \boldsymbol{p}^{n} = \boldsymbol{z}^{n} + \beta^{n}\boldsymbol{p}^{n}$ 
    \ELSE
        \STATE $ \boldsymbol{p}^{n} = \boldsymbol{z}^{n} $   
    \ENDIF
\STATE

$ \boldsymbol{q}^{n} = \texttt{DSE}(\boldsymbol{p}^{n}, \boldsymbol{B},  \boldsymbol{E}_k, \boldsymbol{E}_k^T, \boldsymbol{C}^{-1}, k)$ \Comment{Algorithm~\ref{alg:DSE}} \label{alg:DPCG:MV2}

    \STATE $\alpha^{n} = \rho^{n} / {\boldsymbol{p}^{n}}^T\boldsymbol{q}^{n} $   \label{DPCG:dot2}
    \STATE $\boldsymbol{x} ^ {n+1} = \boldsymbol{x} ^ {n} + \alpha^{n}\boldsymbol{p}^{n}$   
    \STATE $\boldsymbol{r} ^ {n+1} = \boldsymbol{r} ^ {n} - \alpha^{n}\boldsymbol{q}^{n-1}$  
    \STATE $n = n + 1$ 
\ENDWHILE
\RETURN $\boldsymbol{x}^{n}$
\end{algorithmic}
\end{algorithm}

\begin{algorithm}[t!]
\caption{\textbf{Distributed Schur Elimination (DSE)}}
\label{alg:DSE}
\begin{algorithmic}[1]
\REQUIRE Vector $\boldsymbol{x}$, matrix $\boldsymbol{B}, \boldsymbol{E}_k, \boldsymbol{E}_k^T, \boldsymbol{C}^{-1}$, and local GPU rank $k$
\ENSURE Schur elimination result $[\boldsymbol{B} - \boldsymbol{E}\boldsymbol{C}^{-1}\boldsymbol{E}]\boldsymbol{x}$, where $\boldsymbol{E} = \sum_{i=1}^K \boldsymbol{E}_i, \boldsymbol{E}^T = \sum_{i=1}^K \boldsymbol{E}_i^T$
    \STATE $\boldsymbol{a}_{k} = \boldsymbol{E}_{k}^T\boldsymbol{x}$
    \label{alg:DSE:ETx}
	\STATE $\boldsymbol{a} = allreduce(\boldsymbol{a}_k)$
	\Comment{$\sum_{i=1}^K\boldsymbol{a}_i$}\label{alg:DSE:reduce_tau}
	\STATE $ \boldsymbol{b} = \boldsymbol{C}^{-1}\boldsymbol{a} $
	\label{alg:DSE:Ctau}
	\STATE $\boldsymbol{c}_{k} = \boldsymbol{E}_{k}\boldsymbol{b}$\label{alg:DSE:Eeta}
	\STATE $\boldsymbol{c} = allreduce(\boldsymbol{c}_k)$
	\Comment{$\sum_{i=1}^K\boldsymbol{c}_i$}\label{alg:DSE:reduce_gamma}
	\STATE $ \boldsymbol{d} = \boldsymbol{B}\boldsymbol{x} $\label{alg:DSE:Bx}
	\RETURN $\boldsymbol{d} - \boldsymbol{c}$
\end{algorithmic}
\end{algorithm}

\subsection{Distributed PCG}
\label{sec:DPCG}

We then discuss how to distribute the PCG algorithm in BA, shown in Algorithm~\ref{alg:distributed_BA}. This algorithm first constructs a linear system defined in Equation~\ref{eq:schur_complement}. It then solves Equation~\ref{eq:schur_x_c} and computes increment following Equation~\ref{eq:schur_x_p}. It finally uses the increments update state $\boldsymbol{x}$, and tested if a convergence criterion has been met. To guarantee that the distributed BA algorithm achieves the convergence performance, we make  Algorithm~\ref{alg:distributed_BA}, named DPCG, return the same result as the non-distributed linear solver. 


In the following, we describe the execution of DPCG. DPCG takes BA initial state $\boldsymbol{x}^{0}$, matrix block $\boldsymbol{B}$, $\boldsymbol{E}_k$, $\boldsymbol{E}_k^T$, $\boldsymbol{C}$ of $\boldsymbol{H}_k$, constant vector $\boldsymbol{b}$ as input and output solution $\boldsymbol{x}$ for linear system [$\boldsymbol{B} - \boldsymbol{E}\boldsymbol{C}^{-1}\boldsymbol{E}^T] \boldsymbol{x} = \boldsymbol{b}$, where $\boldsymbol{E} = \sum_{k=1}^K\boldsymbol{E}_k$ and $\boldsymbol{E}^T = \sum_{k=1}^K\boldsymbol{E}_k$. The procedures of DPCG using Schur elimination is similar to single-node PCG. Notably, the coefficient matrix of the linear system to be solved is Schur complement. The matrix-vector multiplication operations (Line~\ref{alg:DPCG:MV1},~\ref{alg:DPCG:MV2} in Algorithm~\ref{alg:DPCG}) is thus the multiplication between Schur complement and vector. The difference between distributed compared with non-distributed setting is that DPCG only assign $\boldsymbol{E}_k$ and $\boldsymbol{E}_k^T$ rather than the complete matrices $\boldsymbol{E}$ and $\boldsymbol{E}^T$ to GPU $k$, so we need to guarantee operations that use $\boldsymbol{E}_k$ and $\boldsymbol{E}_k^T$ have the same output compared with using $\boldsymbol{E}$ and $\boldsymbol{E}^T$, these operations happen when doing Schur elimination (Line~\ref{alg:DPCG:MV1},~\ref{alg:DPCG:MV2}).

Our key idea of computing Schur elimination in a distributed manner is that: the summation of matrix-vector multiplication is the same as the the result matrix summation multiplies vector, i.e., $\sum_{k=1}^K(\boldsymbol{E}_k\boldsymbol{v}) = \sum_{k=1}^K(\boldsymbol{E}_k)\boldsymbol{v}$. We compute an intermediate vector  (Line~\ref{alg:DSE:ETx}) and reduce it (Line~\ref{alg:DSE:reduce_tau}), then we compute intermediate vectors sequentially (Line~\ref{alg:DSE:Ctau},~\ref{alg:DSE:Eeta}). We perform all-reduce operation over the intermediate vector (Line~\ref{alg:DSE:reduce_gamma}) and compute another intermediate vector (Line~\ref{alg:DSE:Bx}. After those procedures, we do subtraction to the last two intermediate vectors and output the final result. The result would be the same as computing the complete Schur complement $[\boldsymbol{B} - \boldsymbol{E}\boldsymbol{C}^{-1}\boldsymbol{E}]$ then multiplying it with vector $\boldsymbol{x}$.

\subsection{Complexity Analysis}

In the end, we present the complexity analysis of \sys.
Assume that \sys is given $m$ cameras, $n$ points, and $k$ observations and we often have $k \gg m, n$, the time complexity for building the linear system is $\mathcal{O}(m + n + k)$ and the time complexity for each iteration of the conjugate gradient is $\mathcal{O}(m + n + k)$. Assume we distribute the problem to $K$ GPUs, on each GPU, the time complexity for building the linear system is $\mathcal{O}(m + n + k/K)$ and the time complexity for each iteration of the conjugate gradient is $\mathcal{O}(m + n + k/K)$. The ring all-reduce communication time complexity of each conjugate gradient iteration is $\mathcal{O}(m + n)$. In summary, the total complexity of \sys is $\mathcal{O}(m + n + k/K)$.

%% file: src/5GPU.tex
\section{\sys Implementation}
\label{sec:implementation}

This section describes the implementation of \sys. 
There are several goals of our implementation: (i)~We want to use as many SIMD operations as possible because both computation and memory operations on GPU are essentially SIMD-friendly. (ii)~We want to optimise the memory efficiency of \sys, thus avoiding memory allocation and deallocation; (iii)~We want to implement the APIs of \sys that are fully compatible with existing popular BA libraries: g2o and Ceres. In the following, we highlight how \sys  achieves these implementation goals.

\subsection{SIMD-Friendly Data Structures}
\emph{JetVector} is a novel data structure to perform auto-differentiation over millions of edges. Compared to conventional BA data structure: \emph{Jet} implemented in Ceres, \emph{JetVector} represents a list of Jets (i.e., Array-of-Structure) as a single data object where Jet's data fields: \textit{data} and \textit{grad} across all items are represented as single arrays (i.e., Array-within-Structure). When we perform mathematical operations on \emph{JetVector}, we will start as many GPU threads as the elements in it, every GPU thread process one element. Because \textit{data} and \textit{grad} are stored in the structure of Array-within-Structure, the memory transactions are coalesced and make it easy to reach a high memory throughput. The detailed structure layout of \textit{JecVector} could be found in supplementary materials.

Besides \emph{JetVector}, other parts of \sys are also implemented as SIMD-friendly data structures. The construction of linear system (Line~\ref{alg:distributed_BA:build_linear} in Algorithm~\ref{alg:distributed_BA}) uses L1 cache on GPU to store Jacobian blocks in a SIMD manner. The DPCG algorithm includes a lot of matrix/vector operations which also be benefited from the SIMD structure. A full list of SIMD operations implemented in \sys can be found in supplementary materials.
\label{sec:JV}

\subsection{Memory-Efficient Runtime}

BA computation involves massive objects to be allocated in GPU memory. To avoid expensive memory allocation~\cite{koliousis12crossbow}, we leverage a key observation in BA computation: The automatic differentiation works on GPU buffers that are in the same size across BA iterations. By monitoring the sizes of GPU buffers used in the forward pass of differentiating the BA errors, we can predict the sizes of all memory buffers involved in future BA iterations. Based on this observation, we can pre-allocate these memory buffers in a memory pool, thus avoiding calling the CUDA driver to allocate memory during runtime.

\subsection{Easy-to-use APIs}

The APIs of \sys comprises of two major components:

\noindent
\textbf{(i) Declaring BA problems.} Following the API convention of g2o and Ceres, a BA problem in \sys is declared a graph that contains nodes and edges. The \sys nodes
describe the 3D coordinates or the poses of cameras and these nodes can be directly imported from g2o and Ceres applications. The \sys edges are error functions that can be written using the Eigen library~\cite{eigen}, identical to Ceres.
A \sys user can build a large BA problem by adding BA nodes and edges (using the g2o-equivalent \emph{addEdge} and \emph{addNode} functions).

\noindent
\textbf{(ii) Choosing BA solvers.} \sys also support users to choose solvers given the characteristics of their BA problems. The default solver is the SIMD-optimised DPCG which can automatically use multiple GPUs. Given a small-scale BA problem where intrinsic parallelism is not sufficient, \sys provides users with the CPU-optimised CHOLMOD solver~\cite{CHOLMOD}.

%% file: src/6Experiment.tex
\section{Experimental Evaluation}

\begin{table}[!ht]
\centering
\begin{tabular}{|c|l|l|l|}
\hline
\multicolumn{1}{|l|}{Benchmark}  & \multicolumn{1}{c|}{\textbf{Dataset}} & \#Points & \#Observations \\ \hline
BAL    & Trafalgar-257                         & 65132    & 225911         \\ \cline{2-4} 
                        & Ladybug-1723                          & 156502   & 678718         \\ \cline{2-4} 
                        & Dubrovnik-356                         & 226730   & 1255268        \\ \cline{2-4} 
                        & Venice-1778                           & 993923   & 5001946        \\ \cline{2-4} 
                        & Final-13682                           & 4456117  & 28987644       \\ \hline
Synthesised             & Synthesised-20000                     & 80000    & 80000000       \\ \hline
1DSfM & Alamo-577                             & 140080   & 816891         \\ \cline{2-4} 
                        & Ellis\_Island-233                     & 9210     & 20500          \\ \cline{2-4} 
                        & Gendarmenmarkt-704                    & 76964    & 268747         \\ \cline{2-4} 
                        & Madrid\_Metropolis-347                & 44479    & 195660         \\ \cline{2-4} 
                        & Montreal\_Notre\_Dame-459             & 151876   & 811757         \\ \cline{2-4} 
                        & Notre\_Dame-548                       & 224153   & 1172145        \\ \cline{2-4} 
                        & NYC\_Library-334                      & 54757    & 211614         \\ \cline{2-4} 
                        & Piazza\_del\_Popolo-336               & 29731    & 150161         \\ \cline{2-4} 
                        & Piccadilly-2292                       & 184475   & 798085         \\ \cline{2-4} 
                        & Roman\_Forum-1067                     & 223844   & 1031760        \\ \cline{2-4} 
                        & Tower\_of\_London-484                 & 126648   & 596690         \\ \cline{2-4} 
                        & Trafalgar-5052                        & 327920   & 1266102        \\ \cline{2-4} 
                        & Union\_Square-816                     & 26430    & 90668          \\ \cline{2-4} 
                        & Vienna\_Cathedral-846                 & 154394   & 495940         \\ \cline{2-4} 
                        & Yorkminster-429                       & 100426   & 376980         \\ \hline
\end{tabular}
\caption{\textbf{Dataset Statistics.}}
\label{tab:dataset}
\end{table}

We conduct a comprehensive evaluation with \sys. The evaluation comprises of BAL~\cite{BAL}, 1DSfM~\cite{1dsfm}, and a large synthetic dataset modelled after a city-scale BA application we have in production. The dataset statistic is shown in Table~\ref{tab:dataset}. Due to the page limit, this section only presents the results with BAL~\cite{BAL}, and we put the results of 1DSfM and the synthetic dataset in the supplementary materials.

We compare \sys with four baselines: (i)~Ceres~\cite{ceres} (version 2.0) is the most popular BA library that can efficiently use massive CPU cores,
(ii)~g2o~\cite{g2o} is a lightweight CPU-based BA library,
(iii)~RootBA~\cite{RootBA} is a recent CPU-based BA library that uses Nullspace-Marginalization in place of Schur Complement,
and (iv)~DeepLM~\cite{DeepLM} is the state-of-the-art GPU-based BA library (2021), and it was shown to out-perform other popular BA libraries: STBA~\cite{stba} and PBA~\cite{pba} (We provide comparison results between PBA and MegBA in the supplementary materials).

We run experiments on a server that has 80 Intel Xeon 2.5GHz CPU cores, 8 Nvidia V100 GPUs and 320GB memory. The GPUs are inter-connected using NVLink 2.0. We use 64-bit floating points (FP64) unless otherwise specified. 


 

\input{table/figure}

\input{table/all_time}

\subsection{Overall Performance}

We first evaluate the overall performance of \sys, Ceres, g2o, RootBA, and DeepLM. \sys uses from 1 to 8 GPUs, and CPU-based algorithms use 16 threads. We measure the Mean Squared Error (MSE) in pixels over time.

Figure~\ref{fig:megabbb} shows the evaluation results. In the Venice-1778 dataset (Figure~\ref{fig:megabbb}(a)), \sys achieves the best performance with 8 GPUs, while DeepLM can only use a single GPU. \sys completes with 3.34 seconds while Ceres, RootBA, g2o uses 319.0, 73.94, and 890.6 seconds, respectively. It shows the substantial speed-up (95.5$\times$, 22.1$\times$, and 266.6$\times$), which indicates the benefits of fully exploiting GPUs to accelerate BA computation. For GPU-based BA libraries, \sys can complete with 11.96 seconds while DeepLM spent 24.44 seconds, showing the effectiveness of implementing full vectorisation for BA  on a single GPU. With more GPUs, \sys out-performs DeepLM by 7.316$\times$, which reflects the necessity of adopting multiple GPUs.

Thanks to the vectorisation and distributed BA designs, \sys becomes the state-of-the-art in the large BA dataset (i.e., Final-13682). As shown in Figure~\ref{fig:megabbb}(b), \sys completes in 22.10 seconds, while DeepLM uses 149.6 seconds (6.769$\times$ speed-up), Ceres uses 916 seconds (41.45$\times$ speed-up), g2o uses 13161 seconds (595.5$\times$ speed-up), and RootBA uses 1427 seconds (64.57$\times$ speed-up).  In other datasets (Figure~\ref{fig:megabbb}(c)-(e)), we observe similar speed-up achieved by \sys, indicating the general effectiveness of our proposed approaches. We omit the discussion of these datasets, and their results are reported in Table~\ref{tab:all_time}.

\subsection{Scalability}

\input{table/megba-multicard}

Table~\ref{tab:multi-card} further provides the detailed experimental results to show the scalability of \sys, Ceres and DeepLM. In the Venice-1778 dataset, \sys can consistently improve its performance by adding GPUs (from 11.96 seconds to 3.34 seconds if we increase the number of GPUs from 1 to 8). In addition, the large dataset (Final-13682) can better show the scalability of \sys. By increasing the number of GPUs from 4 to 8, the time can be reduced to 22.10 seconds.

\subsection{Floating point precision}

The accuracy of solving a BA problem is sensitive to the choice of floating point precision (i.e., 32-bit vs. 64-bit floating points). We further evaluate \sys in all datasets with 32-bit and 64-bit floating points, and we report the results of Venice-1778 and Final-13682 in Table~\ref{tab:float}. Other datasets show consistent results and we omit them here. In the dataset of Final-13682, with 4 GPUs, \sys (FP32) can complete in 4.804 seconds and \sys (FP64) can complete in 28.70 seconds, while both of them are reaching the same MSE. 
This shows the exactness of the distributed BA algorithm in \sys. Even with lower precision, \sys can reach the same MSE as double-precision; but offering 5.97$\times$ speed up, making \sys (FP32) be the state-of-the-art in Final-13682.

\input{table/megba-float}

%% file: table/figure.tex
\begin{figure}[t!]
\centering
\subcaptionbox{Venice-1778}{\includegraphics[width=0.49\textwidth]{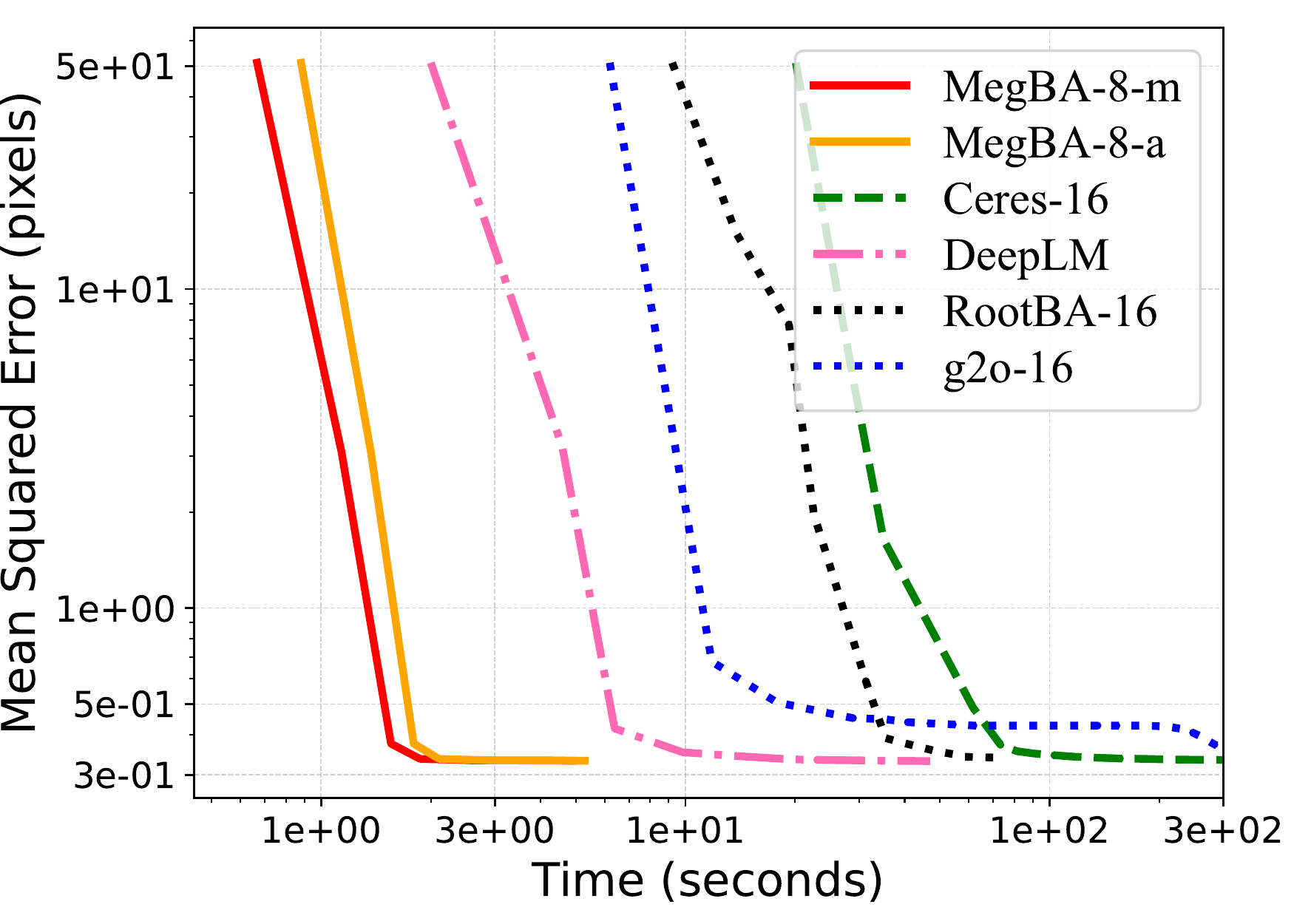}}%
\hfill
\subcaptionbox{Final-13682}{\includegraphics[width=0.49\textwidth]{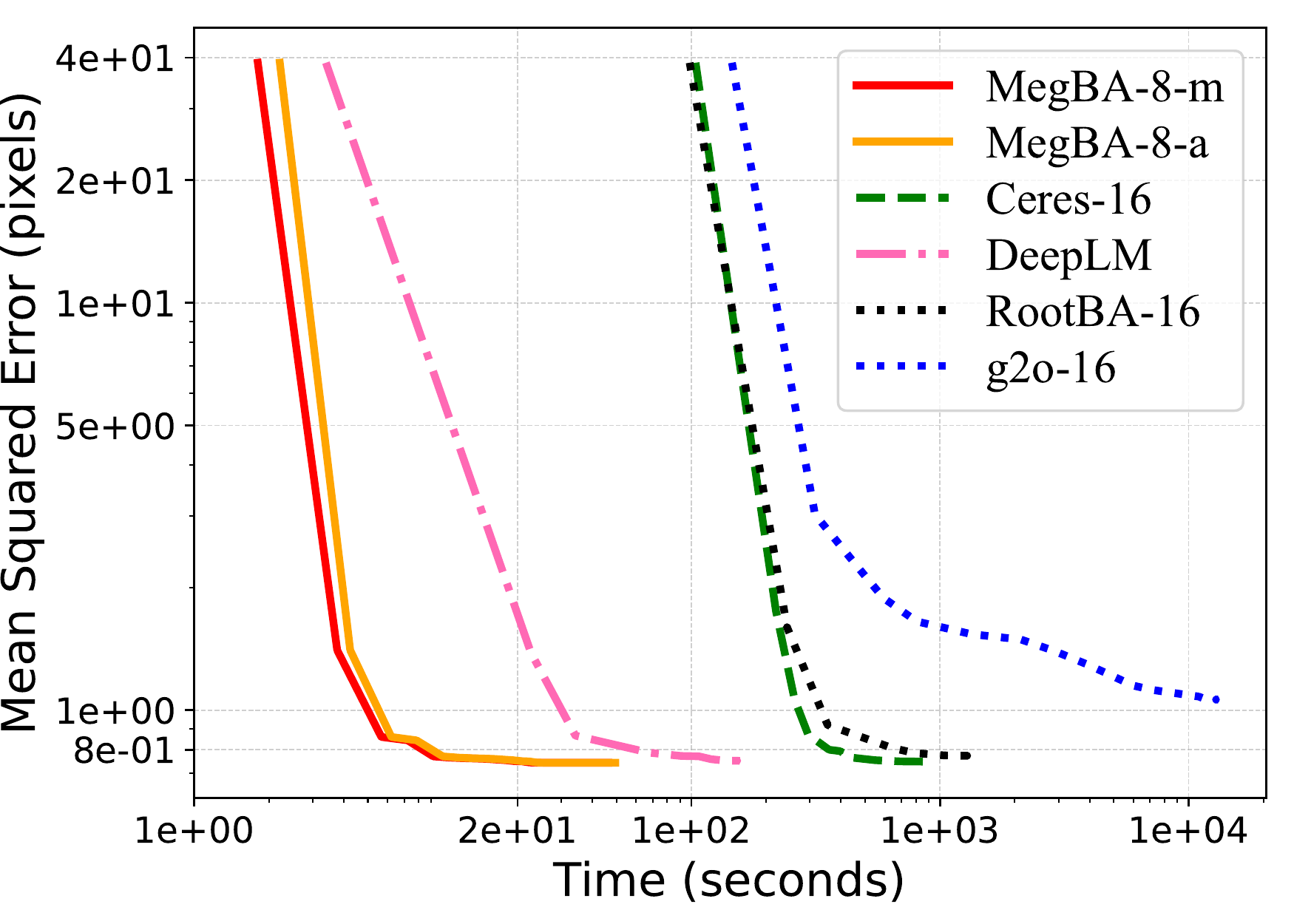}}%
\hfill
\subcaptionbox{Ladybug-1723}{\includegraphics[width=0.32\textwidth]{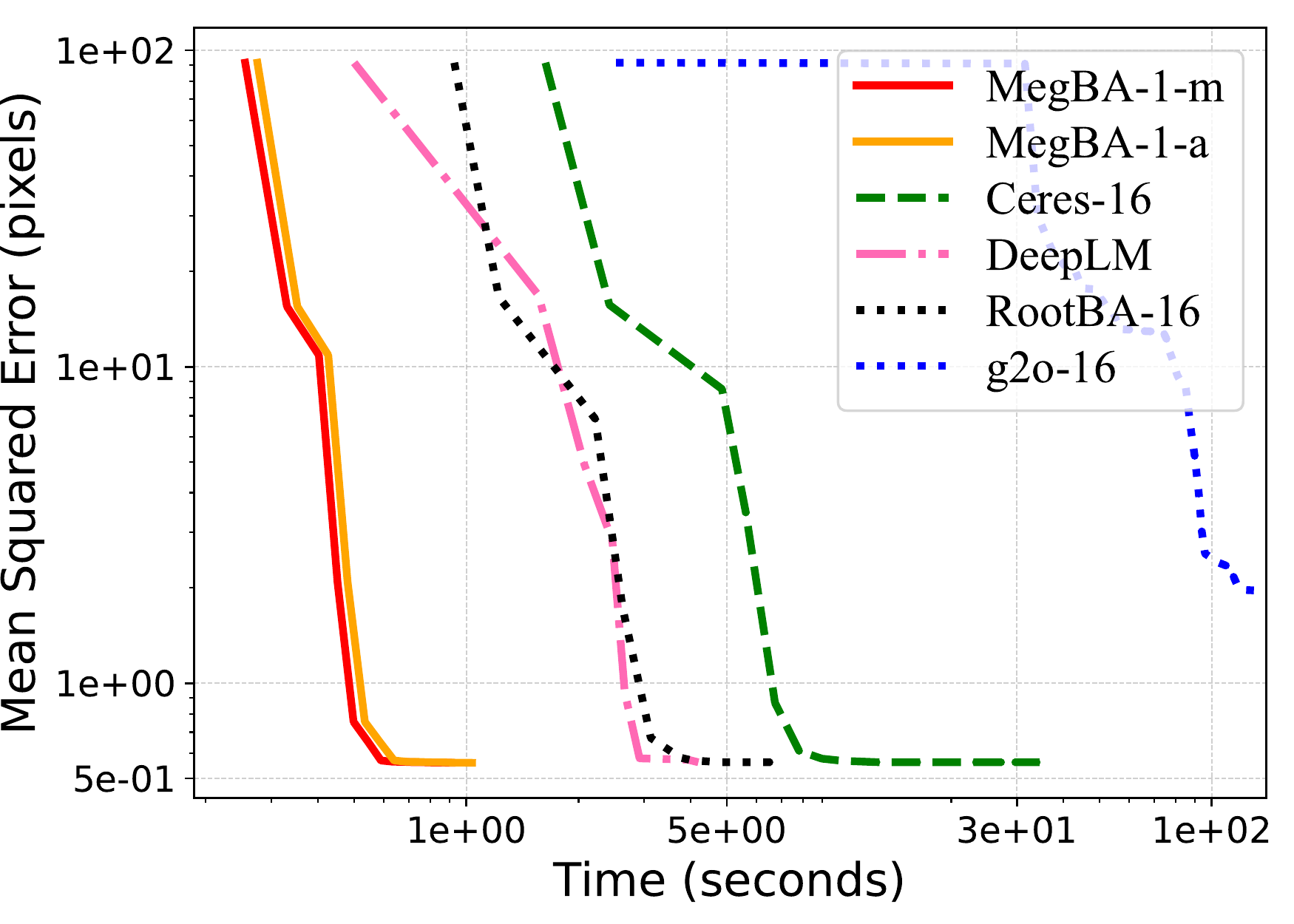}}%
\hfill
\subcaptionbox{Trafalgar-257}{\includegraphics[width=0.32\textwidth]{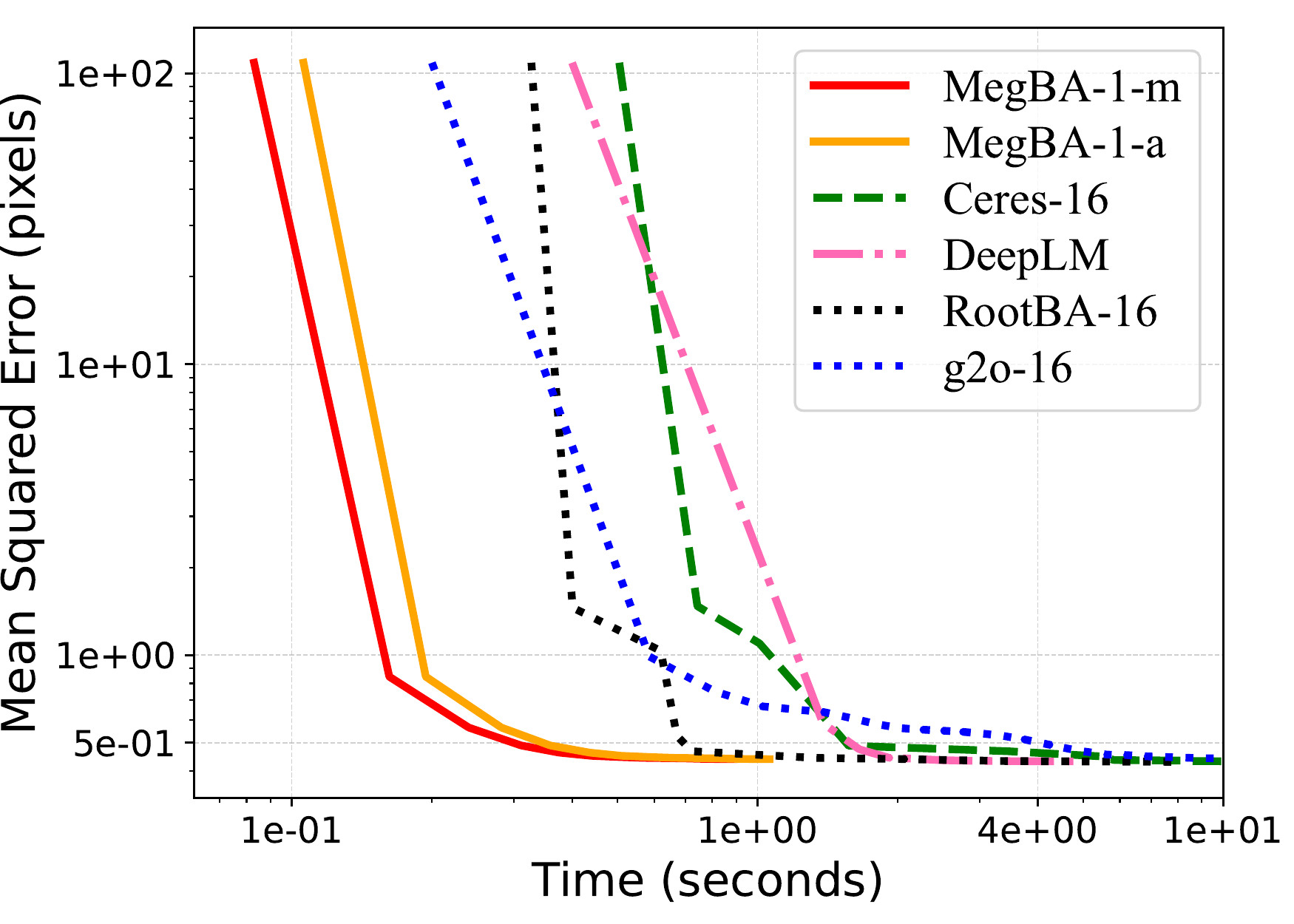}}
\hfill
\subcaptionbox{Dubrovnik-356}{\includegraphics[width=0.32\textwidth]{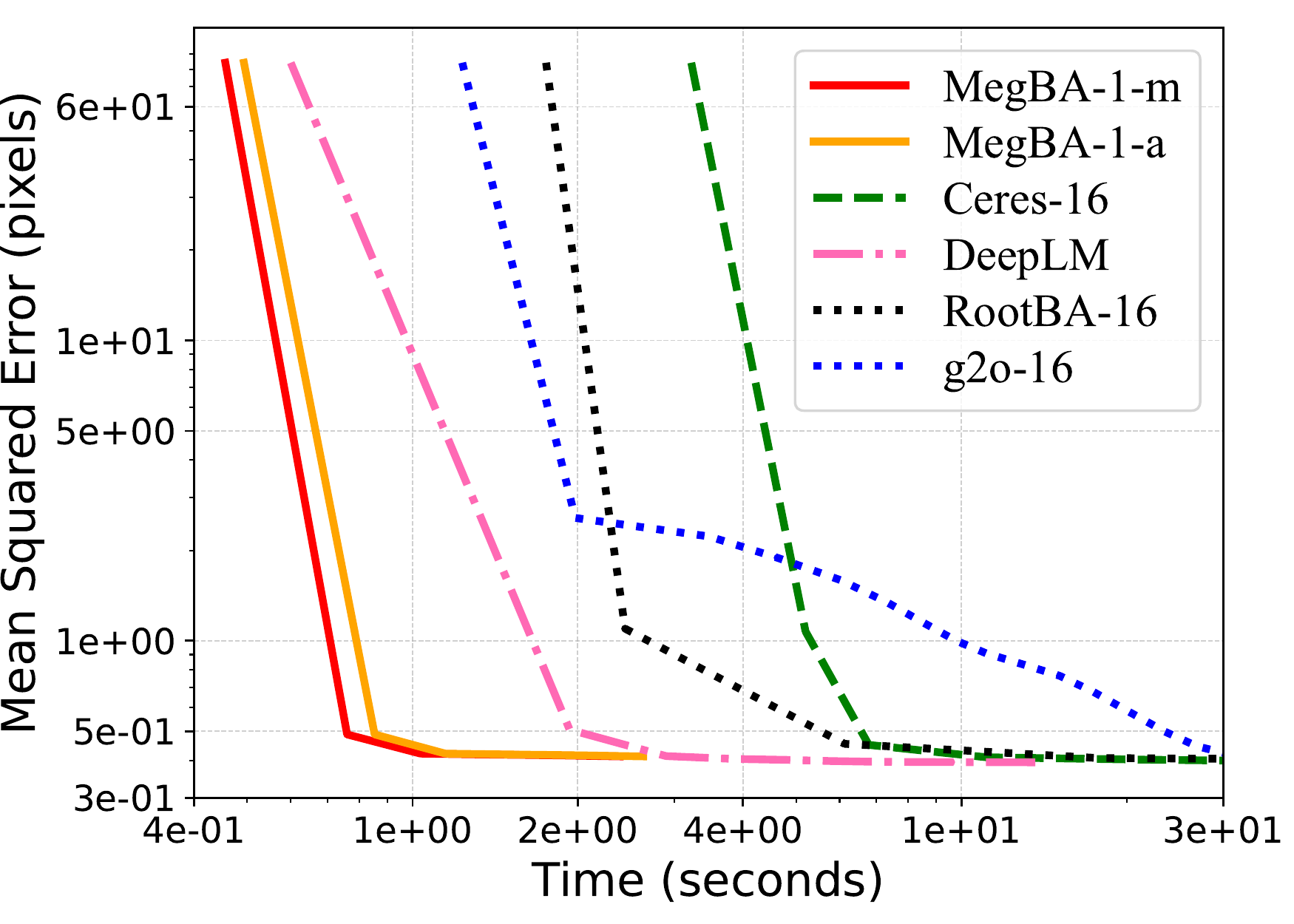}}
\hfill

\caption{\textbf{Mean Squared Error over Time.} \textit{MegBA-$X$-Y refers to $X$ GPUs while \textbf{-a} refers to auto-differentiation Jacobian and \textbf{-m} refers to analytical differentiation Jacobian. Ceres/RootBA/g2o-$X$ refers to $X$ CPU threads.}}

\label{fig:megabbb}
\end{figure}

%% file: table/all_time.tex
\begin{table*}[ht!]
\centering
\begin{tabular}{|l|lll|lll|lll|}
\hline
         & \multicolumn{3}{c|}{Trafalgar-257}                                                & \multicolumn{3}{c|}{Ladybug-1723}                                                & \multicolumn{3}{c|}{Dubrovnik-356}                                              \\ \hline
         & \multicolumn{1}{l|}{MSE}            & \multicolumn{1}{l|}{Time}          & Mem   & \multicolumn{1}{l|}{MSE}            & \multicolumn{1}{l|}{Time}          & Mem   & \multicolumn{1}{l|}{MSE}            & \multicolumn{1}{l|}{Time}          & Mem  \\ \hline
Ceres-16           & \multicolumn{1}{l|}{0.434}          & \multicolumn{1}{l|}{8.160}          & 1.659                    & \multicolumn{1}{l|}{0.562}          & \multicolumn{1}{l|}{34.50}         & 2.093                    & \multicolumn{1}{l|}{0.393}          & \multicolumn{1}{l|}{116.0}        & 2.550                    \\ \hline
DeepLM             & \multicolumn{1}{l|}{0.434}          & \multicolumn{1}{l|}{3.820}          & 1.445                    & \multicolumn{1}{l|}{0.573}          & \multicolumn{1}{l|}{3.930}          & 2.144                    & \multicolumn{1}{l|}{0.396}          & \multicolumn{1}{l|}{6.119}          & 2.693                    \\ \hline
g2o-16             & \multicolumn{1}{l|}{0.434}          & \multicolumn{1}{l|}{21.69}         & 1.358                    & \multicolumn{1}{l|}{1.961}          & \multicolumn{1}{l|}{140.7}        & 1.866                    & \multicolumn{1}{l|}{0.394}          & \multicolumn{1}{l|}{94.39}         & 2.308                    \\ \hline
RootBA-16          & \multicolumn{1}{l|}{\textbf{0.433}} & \multicolumn{1}{l|}{3.307}          & 1.468                    & \multicolumn{1}{l|}{0.562}          & \multicolumn{1}{l|}{7.050}          & 2.423                    & \multicolumn{1}{l|}{0.393}          & \multicolumn{1}{l|}{78.16}         & 3.942                    \\ \hline
MegBA-1-a            & \multicolumn{1}{l|}{0.438}          & \multicolumn{1}{l|}{1.364}          & 1.270                    & \multicolumn{1}{l|}{0.560}          & \multicolumn{1}{l|}{0.932}          & 2.450                    & \multicolumn{1}{l|}{0.411}          & \multicolumn{1}{l|}{3.640}          & 3.940                    \\ \hline
MegBA-1-m & \multicolumn{1}{l|}{0.438}          & \multicolumn{1}{l|}{\textbf{1.148}} & 1.010                    & \multicolumn{1}{l|}{\textbf{0.560}} & \multicolumn{1}{l|}{\textbf{0.774}} & 1.660                    & \multicolumn{1}{l|}{0.411}          & \multicolumn{1}{l|}{\textbf{3.263}} & 2.480                    \\ \hline

\end{tabular}
\caption{\textbf{Small-scale experiments} \textit{We only report the results of \sys with a GPU because the datasets in this table are small. MSE is the final Mean Squared Error (pixels), Time is BA duration, and Mem is the memory in GB.}}

\label{tab:all_time}
\end{table*}

%% file: table/megba-multicard.tex
\begin{table}[t!]
\centering
\setlength{\tabcolsep}{1.5mm}{
\begin{tabular}{|l|lll|lll|}
\hline
               & \multicolumn{3}{c|}{Venice-1778}                                                      & \multicolumn{3}{c|}{Final-13682}                                                      \\ \hline
               & \multicolumn{1}{l|}{MSE}            & \multicolumn{1}{l|}{Time}           & Mem  & \multicolumn{1}{l|}{MSE}            & \multicolumn{1}{l|}{Time}          & Mem  \\ \hline
Ceres-16           & \multicolumn{1}{l|}{0.334}          & \multicolumn{1}{l|}{319.0}            & 5.983                    & \multicolumn{1}{l|}{0.749}          & \multicolumn{1}{l|}{916.0}             & 26.08                   \\ \hline
DeepLM             & \multicolumn{1}{l|}{0.333}          & \multicolumn{1}{l|}{24.44}         & 6.256                    & \multicolumn{1}{l|}{0.751}          & \multicolumn{1}{l|}{149.6}         & 14.89                   \\ \hline
g2o-16             & \multicolumn{1}{l|}{0.335}          & \multicolumn{1}{l|}{890.6}        & 5.999                    & \multicolumn{1}{l|}{1.061}               & \multicolumn{1}{l|}{13161}                & 36.89                   \\ \hline
RootBA-16          & \multicolumn{1}{l|}{0.337}          & \multicolumn{1}{l|}{73.94}         & 14.14                   & \multicolumn{1}{l|}{0.773}          & \multicolumn{1}{l|}{1,427}           & 263.2                   \\ \hline
MegBA-1-a            & \multicolumn{1}{l|}{0.333}          & \multicolumn{1}{l|}{11.96}         & 13.68                   & \multicolumn{1}{l|}{OOM}            & \multicolumn{1}{l|}{OOM}             & OOM                      \\ \hline
MegBA-2-a            & \multicolumn{1}{l|}{0.333}          & \multicolumn{1}{l|}{7.133}          & 14.51                   & \multicolumn{1}{l|}{OOM}            & \multicolumn{1}{l|}{OOM}             & OOM                      \\ \hline
MegBA-4-a            & \multicolumn{1}{l|}{0.333}          & \multicolumn{1}{l|}{4.767}          & 16.76                   & \multicolumn{1}{l|}{0.748}          & \multicolumn{1}{l|}{28.70}          & 81.03                   \\ \hline
MegBA-8-a            & \multicolumn{1}{l|}{0.333}          & \multicolumn{1}{l|}{3.340}          & 22.61                   & \multicolumn{1}{l|}{0.748}          & \multicolumn{1}{l|}{22.10}          & 89.74                   \\ \hline
MegBA-1-m & \multicolumn{1}{l|}{0.333}          & \multicolumn{1}{l|}{10.92}         & 7.870                    & \multicolumn{1}{l|}{OOM}            & \multicolumn{1}{l|}{OOM}             & OOM                      \\ \hline
MegBA-2-m & \multicolumn{1}{l|}{0.333}          & \multicolumn{1}{l|}{6.618}          & 8.693                    & \multicolumn{1}{l|}{0.748}          & \multicolumn{1}{l|}{50.57}          & 43.60                   \\ \hline
MegBA-4-m & \multicolumn{1}{l|}{0.333}          & \multicolumn{1}{l|}{4.617}          & 10.95                   & \multicolumn{1}{l|}{0.748}          & \multicolumn{1}{l|}{26.46}          & 47.33                   \\ \hline
MegBA-8-m & \multicolumn{1}{l|}{\textbf{0.333}} & \multicolumn{1}{l|}{\textbf{3.014}} & 16.79                   & \multicolumn{1}{l|}{\textbf{0.748}} & \multicolumn{1}{l|}{\textbf{20.68}} & 56.06                   \\ \hline

\end{tabular}}

\caption{\textbf{Large-scale experiments.}}
\label{tab:multi-card}
\end{table}

%% file: table/megba-float.tex
\begin{table}[t!]
\centering
\setlength{\tabcolsep}{0.8mm}{
\begin{tabular}{|l|lll|lll|}
\hline
               & \multicolumn{3}{c|}{Venice-1778}                                                      & \multicolumn{3}{c|}{Final-13682}                                                      \\ \hline
               & \multicolumn{1}{l|}{MSE}            & \multicolumn{1}{l|}{Time}           & Mem  & \multicolumn{1}{l|}{MSE}            & \multicolumn{1}{l|}{Time}          & Mem  \\ \hline
MegBA-1-a(FP32) & \multicolumn{1}{l|}{0.334} & \multicolumn{1}{l|}{2.620}  & 8.300  & \multicolumn{1}{l|}{OOM}   & \multicolumn{1}{l|}{OOM}    & OOM    \\ \hline
MegBA-1-a(FP64) & \multicolumn{1}{l|}{0.333} & \multicolumn{1}{l|}{11.96} & 13.68 & \multicolumn{1}{l|}{OOM}   & \multicolumn{1}{l|}{OOM}    & OOM    \\ \hline
MegBA-1-m(FP32) & \multicolumn{1}{l|}{0.333} & \multicolumn{1}{l|}{2.065}  & 4.821  & \multicolumn{1}{l|}{0.750} & \multicolumn{1}{l|}{11.82} & 24.51 \\ \hline
MegBA-1-m(FP64) & \multicolumn{1}{l|}{0.333} & \multicolumn{1}{l|}{10.92} & 7.870  & \multicolumn{1}{l|}{OOM}   & \multicolumn{1}{l|}{OOM}    & OOM    \\ \hline
MegBA-2-a(FP32) & \multicolumn{1}{l|}{0.333} & \multicolumn{1}{l|}{1.903}  & 8.447  & \multicolumn{1}{l|}{0.750} & \multicolumn{1}{l|}{11.04} & 42.48 \\ \hline
MegBA-2-a(FP64) & \multicolumn{1}{l|}{0.333} & \multicolumn{1}{l|}{7.133}  & 14.51 & \multicolumn{1}{l|}{OOM}   & \multicolumn{1}{l|}{OOM}    & OOM    \\ \hline
MegBA-2-m(FP32) & \multicolumn{1}{l|}{0.333} & \multicolumn{1}{l|}{1.353}  & 5.541  & \multicolumn{1}{l|}{0.750} & \multicolumn{1}{l|}{5.133}  & 25.63 \\ \hline
MegBA-2-m(FP64) & \multicolumn{1}{l|}{0.333} & \multicolumn{1}{l|}{6.618}  & 8.693  & \multicolumn{1}{l|}{0.748} & \multicolumn{1}{l|}{50.57} & 43.60 \\ \hline
MegBA-4-a(FP32) & \multicolumn{1}{l|}{0.333} & \multicolumn{1}{l|}{1.680}  & 10.50 & \multicolumn{1}{l|}{0.749} & \multicolumn{1}{l|}{4.804}  & 45.28 \\ \hline
MegBA-4-a(FP64) & \multicolumn{1}{l|}{0.333} & \multicolumn{1}{l|}{4.767}  & 16.76 & \multicolumn{1}{l|}{0.748} & \multicolumn{1}{l|}{28.70} & 81.03 \\ \hline
MegBA-4-m(FP32) & \multicolumn{1}{l|}{0.334} & \multicolumn{1}{l|}{1.274}  & 7.598  & \multicolumn{1}{l|}{0.748} & \multicolumn{1}{l|}{\textbf{4.279}}  & 28.43 \\ \hline
MegBA-4-m(FP64) & \multicolumn{1}{l|}{0.333} & \multicolumn{1}{l|}{4.617}  & 10.95 & \multicolumn{1}{l|}{0.748} & \multicolumn{1}{l|}{26.46} & 47.33 \\ \hline
MegBA-8-a(FP32) & \multicolumn{1}{l|}{0.334} & \multicolumn{1}{l|}{1.622}  & 16.02 & \multicolumn{1}{l|}{0.748} & \multicolumn{1}{l|}{8.973}  & 52.15 \\ \hline
MegBA-8-a(FP64) & \multicolumn{1}{l|}{0.333} & \multicolumn{1}{l|}{3.340}  & 22.60 & \multicolumn{1}{l|}{0.748} & \multicolumn{1}{l|}{22.10} & 89.74 \\ \hline
MegBA-8-m(FP32) & \multicolumn{1}{l|}{\textbf{0.333}} & \multicolumn{1}{l|}{\textbf{1.271}}  & 12.99 & \multicolumn{1}{l|}{\textbf{0.747}} & \multicolumn{1}{l|}{7.582}  & 35.31 \\ \hline
MegBA-8-m(FP64) & \multicolumn{1}{l|}{0.333} & \multicolumn{1}{l|}{3.014}  & 16.79 & \multicolumn{1}{l|}{0.748} & \multicolumn{1}{l|}{20.68} & 56.06 \\ \hline

\end{tabular}}

\caption{\textbf{Performance with 32-bit and 64-bit floating points.}}
\label{tab:float}
\end{table}

%% file: supp.tex
\appendix

\section{JetVector Design}
To show the novel design of \textit{JetVector}, we compare it with its predecessor: \textit{Jet} implemented in prior BA libraries such as Ceres~\cite{ceres}. 

A \textit{Jet} object comprises \textit{value} and \textit{grad} (i.e., gradients). It has atomic operators such as $+$, $-$, $*$, and $/$ which are essential for auto-differentiation. Multiple Jet objects are contained in an Array-of-Structures (AoS). Though effective for CPUs, the use of AoS makes it difficult to coalesce GPU memory transactions, resulting in low efficiency in utilising GPUs.

\begin{figure}[!ht]
		\label{fig:example}
		\begin{subfigure}{.5\textwidth}
			\centering
			\includegraphics[width=\textwidth]{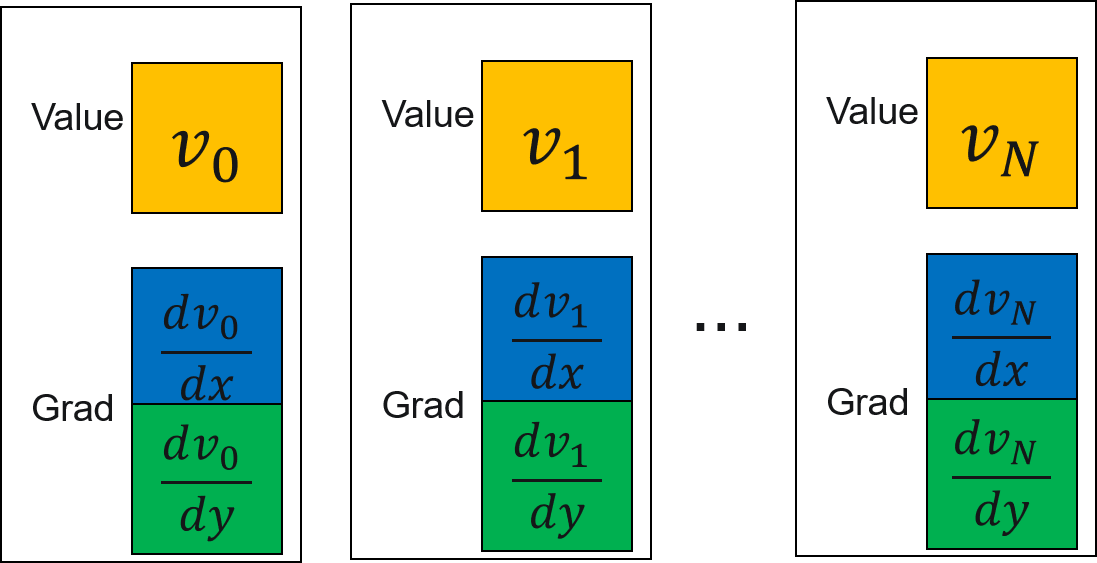}
			\caption{Data structure of \textit{Jet}}
		\end{subfigure}
		\begin{subfigure}{.5\textwidth}
			\centering
			\includegraphics[width=.7\textwidth]{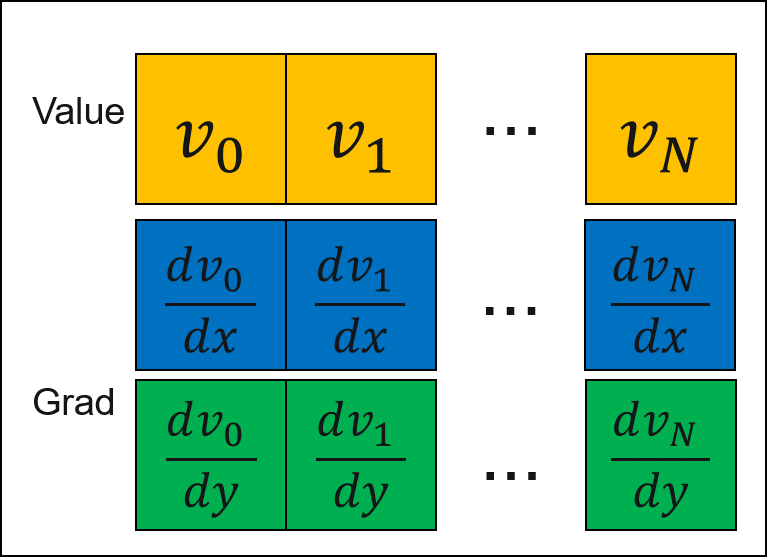}
			\caption{Data structure of \textit{JetVector}}
		\end{subfigure}
		\caption{Comparison between \textit{Jet} and \textit{JetVector}.}
		\label{fig:JetVSJetVector}
\end{figure}

In contrast, \textit{JetVector} is implemented as a Structure-of-Arrays which allows GPU memory transactions to be effectively coalesced. Figure~\ref{fig:JetVSJetVector} compares Jet and JetVector. As we can see, if we stack \textit{Jet} in a vector, the memory addresses of \textit{Value} and \textit{Grad} of different each \textit{Jet} objects are not continuous. In \textit{JetVector}, however \textit{Value} and \textit{Grad} objects are organised in a continuous manner in the memory, making it friendly for GPUs.

\begin{figure}[!ht]
    \centering
    \includegraphics[width=\textwidth]{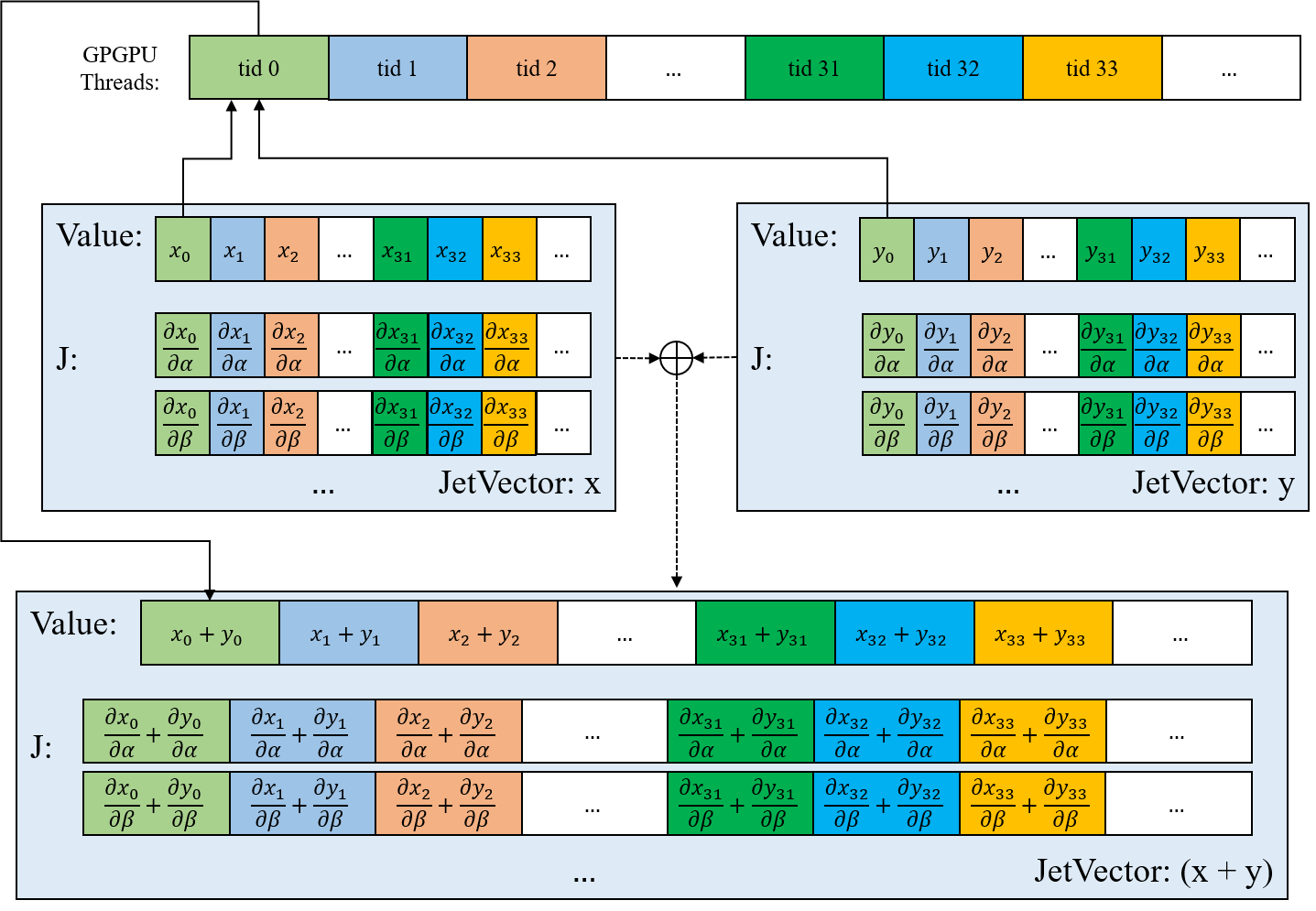}
    \caption{Example of adding two \textit{JetVector} objects.}
    \label{fig:JetVectorAdd}
\end{figure}

We show how does \sys add JetVector objects in Figure~\ref{fig:JetVectorAdd}. Supposing there are $N$ \textit{Jet}s in \textit{JetVector}. We launch $N$ GPU threads and each GPU thread processes a \textit{Jet} object. In this way, GPU threads will access adjacent data elements, thus coalescing memory transactions. As a result, the memory loading/storing operations in a warp (a warp is 32 consecutive GPU threads) can be achieved by a 128-byte memory transaction. Notably, the memory transactions of a naive AoS structure comprises multiple serialised 32-byte memory transactions, which can adversely affect memory performance.

\section{Edge-based Partitioning of BA Problems}

We provide an example that shows how \sys partitions non-zero Hessian blocks and assigns the partitions to multiple GPUs in a load balancing manner.

\begin{figure}[!t]
		\label{fig:example}
		\begin{subfigure}{.5\textwidth}
			\centering
			\includegraphics[width=\textwidth, height=2.6in]{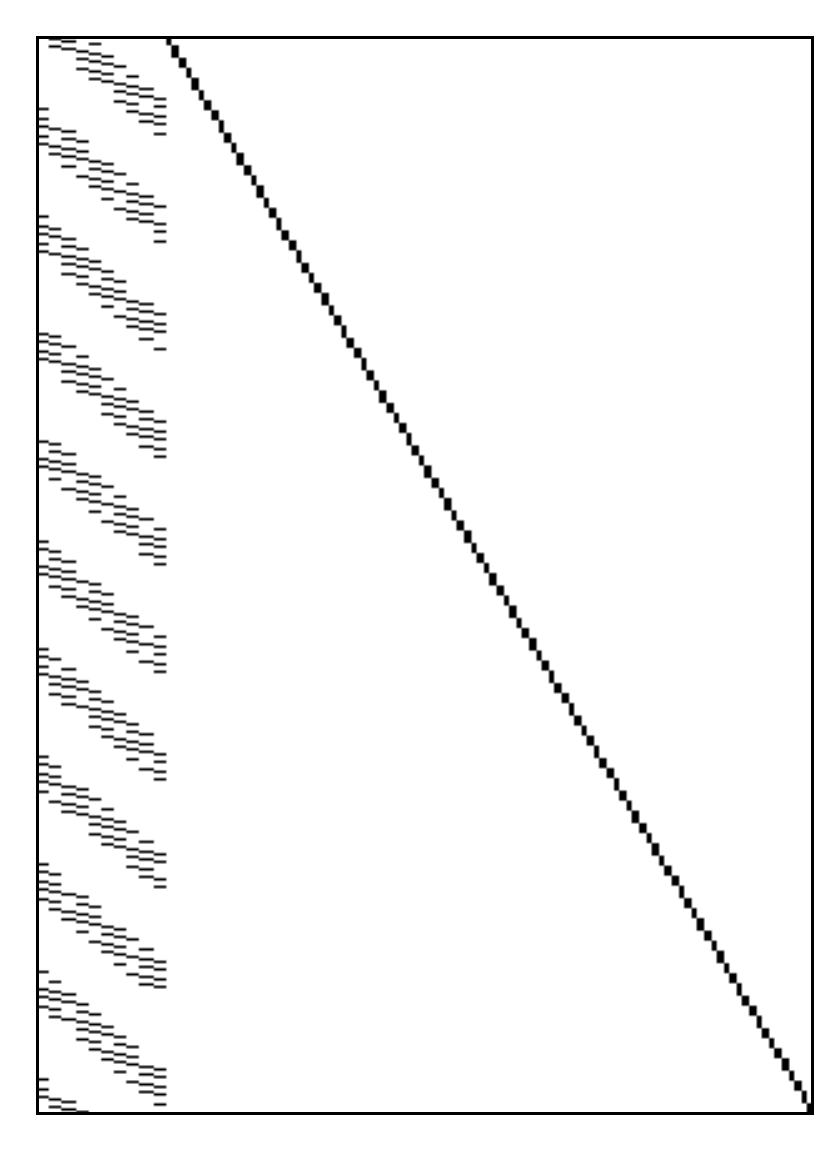}
			\caption{Jacobian $J$}
			\label{fig:Jacobian:J}
		\end{subfigure}
		\begin{subfigure}{.5\textwidth}
			\centering
			\includegraphics[width=.85\textwidth, height=1.1in]{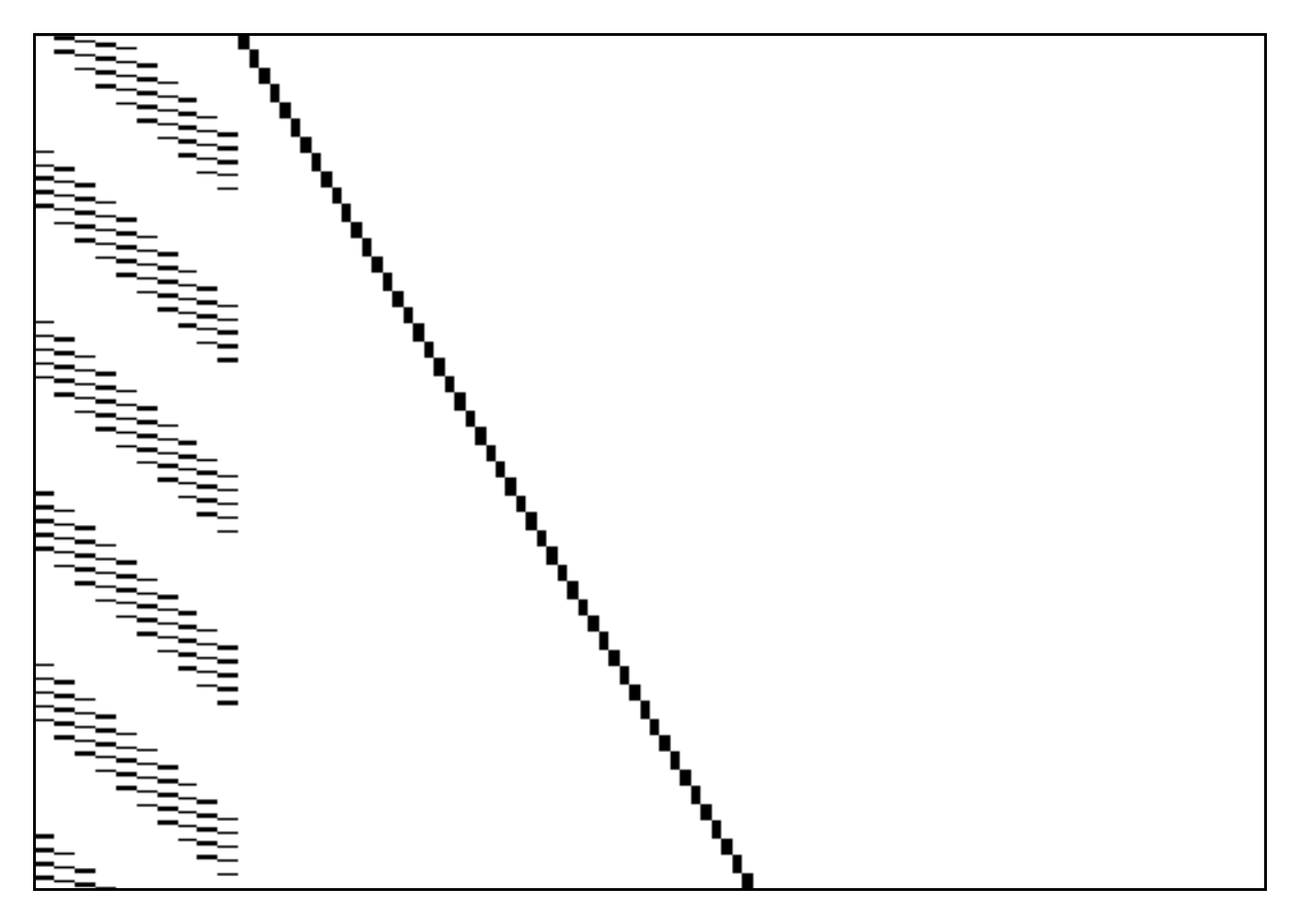}
			\caption{Partitioned Jacobian $J_0$ on GPU $0$}
			\label{fig:Jacobian:J0}
			\includegraphics[width=.85\textwidth, height=1.1in]{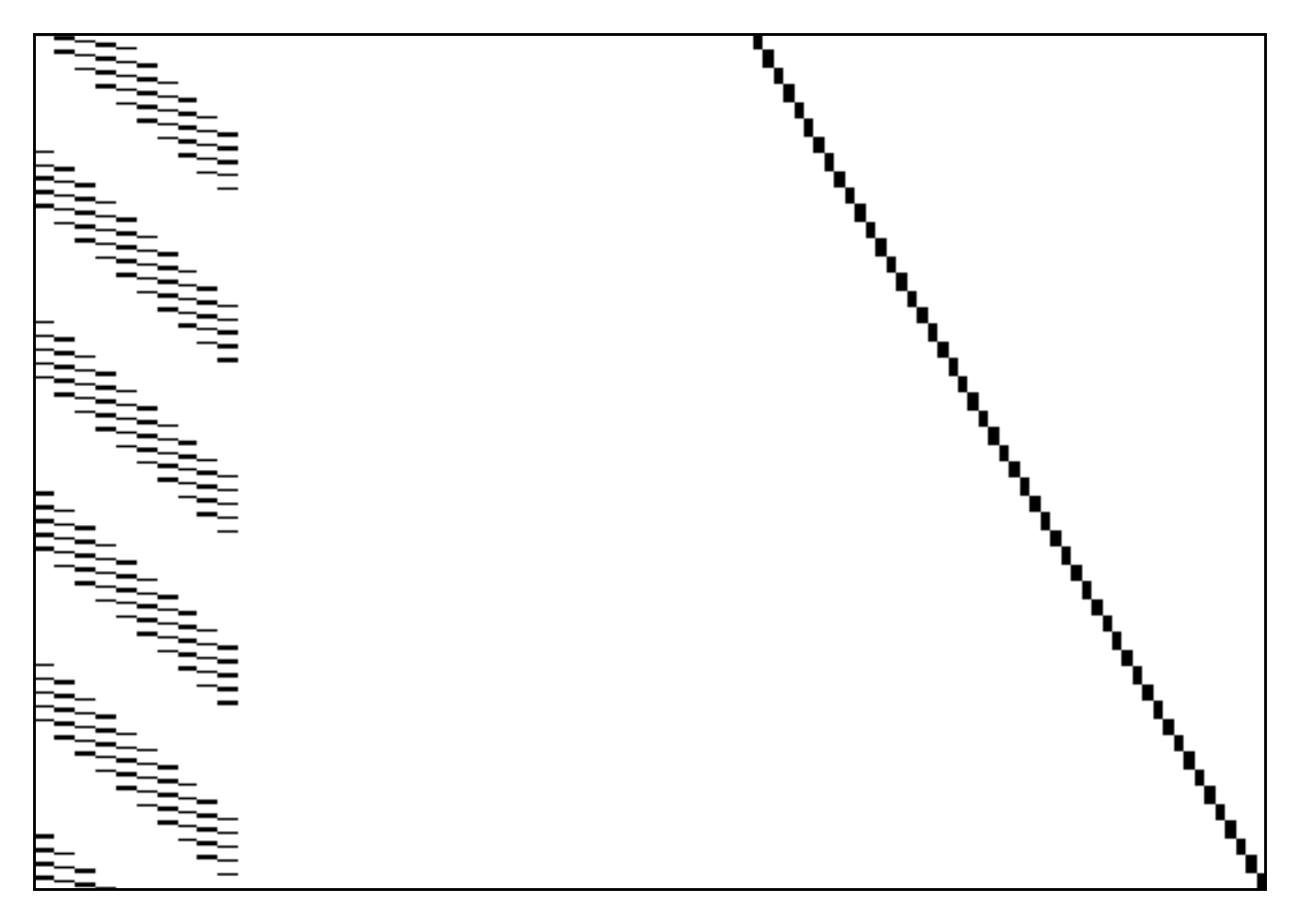}
			\caption{Partitioned Jacobian $J_1$ on GPU $1$}
			\label{fig:Jacobian:J1}
		\end{subfigure}
		\caption{A Jacobian matrix. (a) is the Jacobian of a BA problem. We partition Jacobian into two matrix blocks (b) and (c). \sys stores these two matrix blocks on two GPUs.}
		\label{fig:Jacobian}
\end{figure}

Figure~\ref{fig:Jacobian:J} shows the Jacobian structure of a BA problem. The black blocks refer to non-zero blocks and the white blocks refer to zero blocks. It is partitioned into two sub-matrices shown in Figure~\ref{fig:Jacobian:J0} and Figure~\ref{fig:Jacobian:J1}.

\begin{figure}[!t]
		\label{fig:example}
		\begin{subfigure}{.5\textwidth}
			\centering
			\includegraphics[width=2in, height=2in]{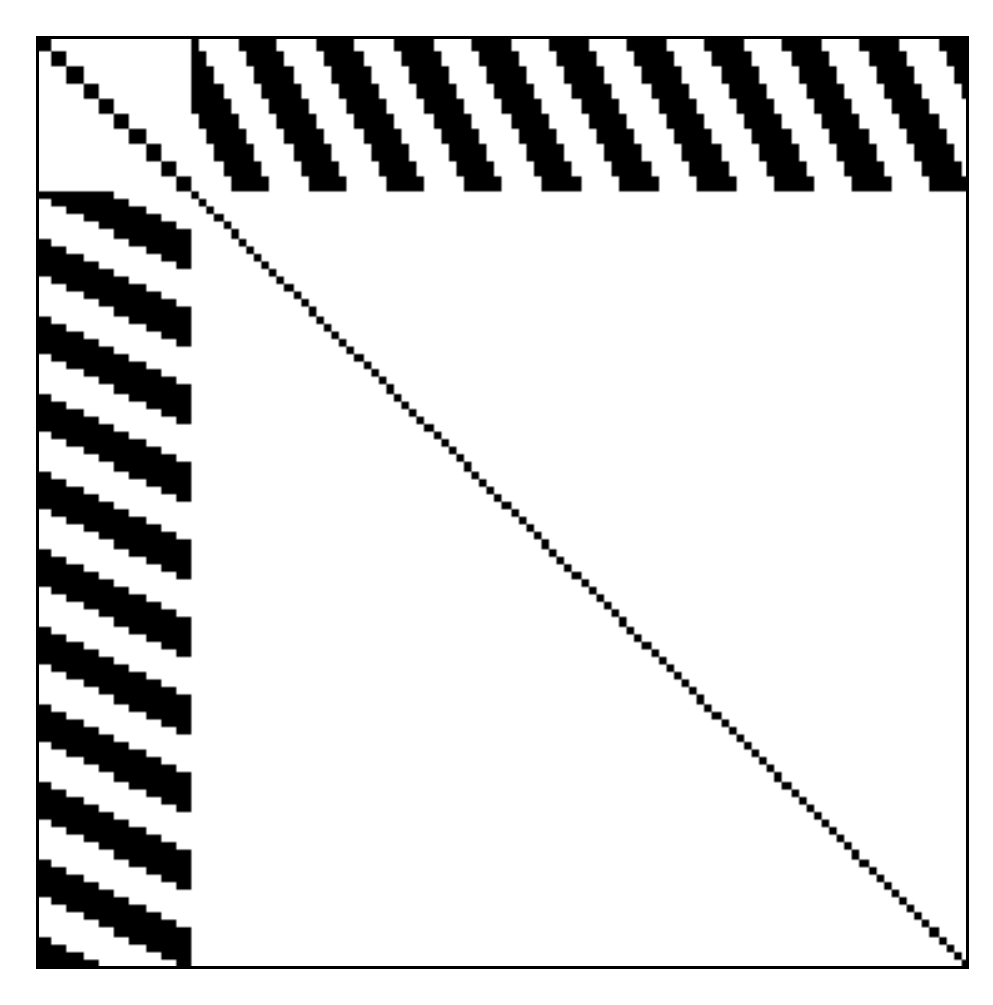}
			\caption{Hessian $H$}
			\label{fig:Hessian:H}
		\end{subfigure}
		\begin{subfigure}{.5\textwidth}
			\centering
			\includegraphics[width=1in, height=1in]{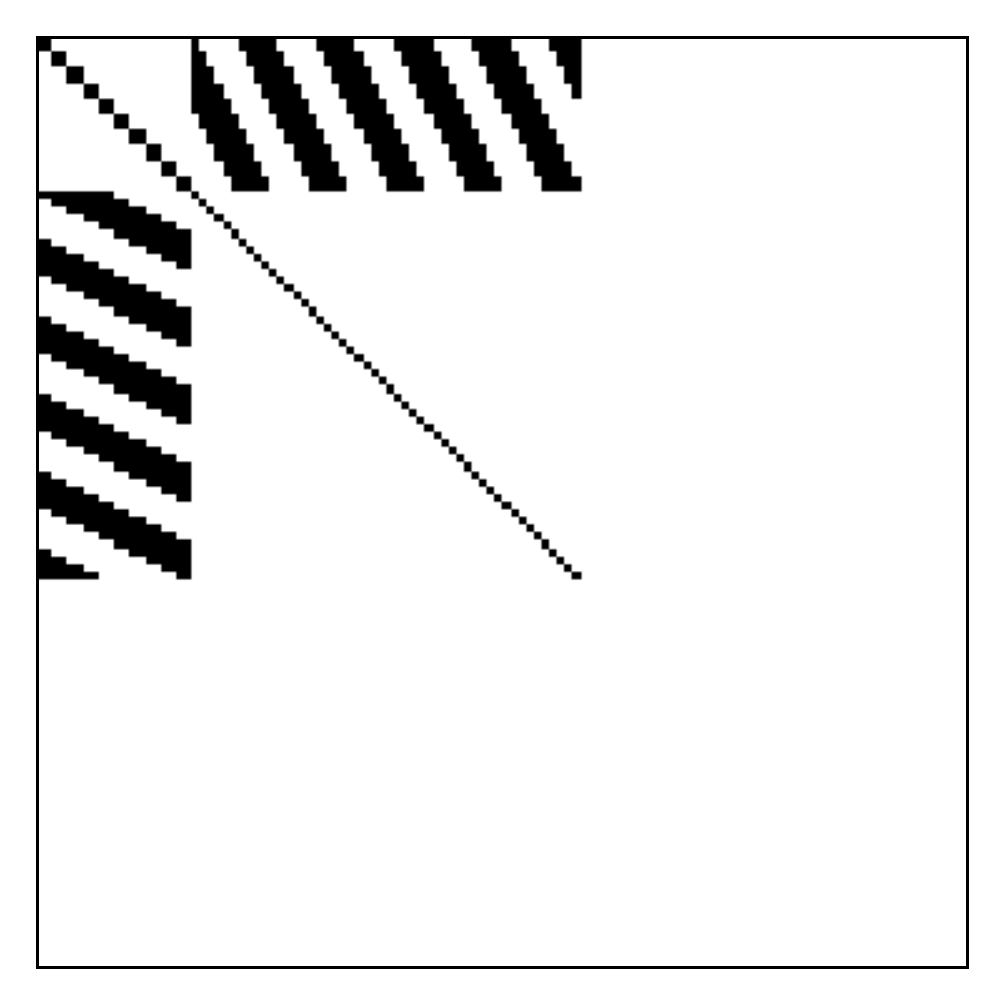}
			\caption{Hessian $H_0$ on GPU $0$}
			\label{fig:Hessian:H0}
			\includegraphics[width=1in, height=1in]{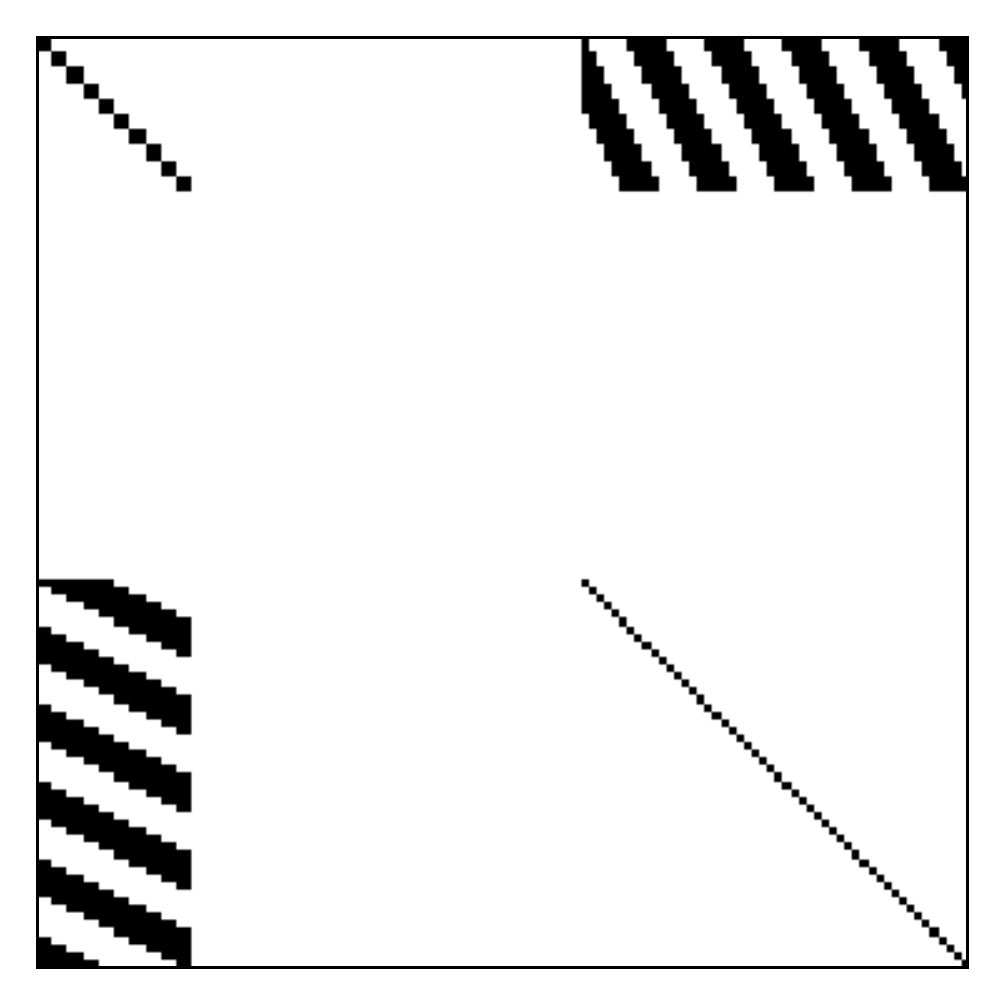}
			\caption{Hessian $H_1$ on GPU $1$}
			\label{fig:Hessian:H1}
		\end{subfigure}
		\caption{A Hessian matrix. (a) is the Hessian $H$ constructed from Jacobian $J$ that $H = J^TJ$. On each GPU, \sys constructs a Hessian matrix so that $H_0 = J_0^TJ_0, H_1 = J_1^TJ_1$.}
		\label{fig:Hessian}
\end{figure}

Figure~\ref{fig:Hessian:H} is the structure of the Hessian matrix to be computed by \sys. A Hessian, though stored in the sparse format, is still too large to fit into a GPU. To address this, \sys stores a part of a Hessian in each GPU, as shown in Figure~\ref{fig:Hessian:H0} and Figure~\ref{fig:Hessian:H1}.
Since we have $H = H_0 + H_1$, it is guaranteed that computations based on partitioned Hessian sub-matrices are equivalent to the original computation.

\input{experiments}

\section{GPU Utilisation}
\begin{figure}[!t]
    \centering
    \includegraphics[width=\textwidth]{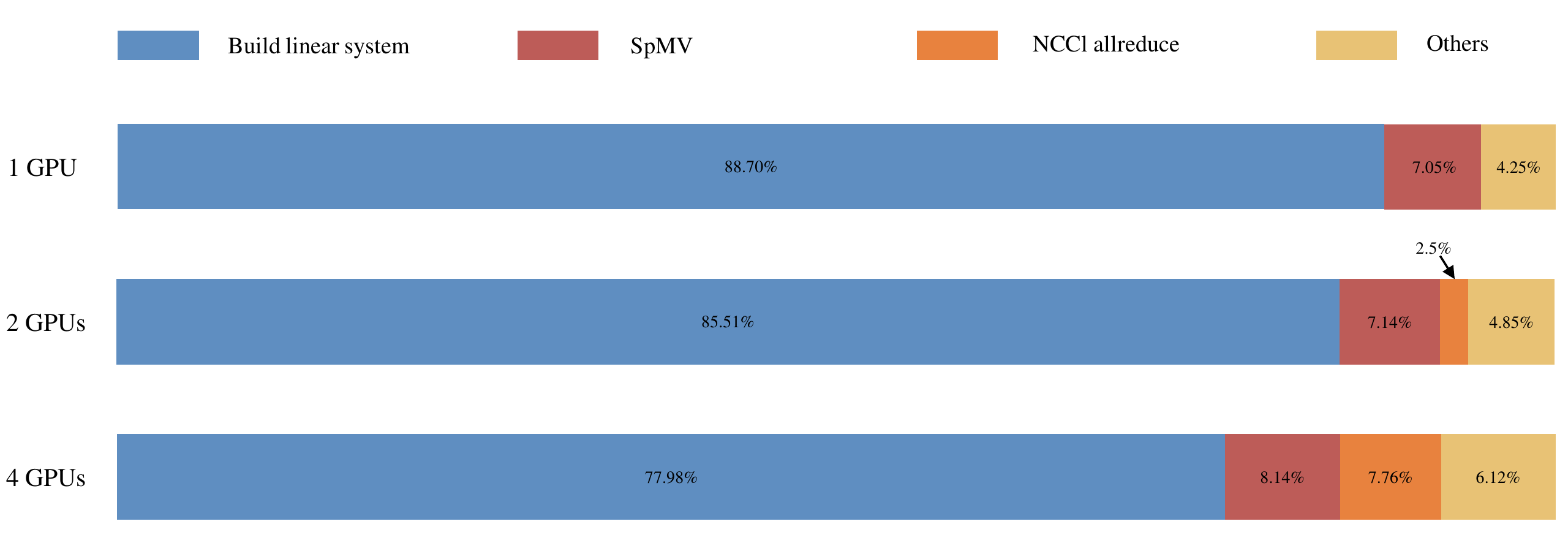}
    \caption{The visualisation of the profiling result returned by nvprof. We use the dataset Venice-1778. SpMV denotes sparse matrix-vector multiplication. Using 2 GPUs, the all-reduce communication accounts for 2.5\%, while the overall time reduces 60.6\% compared with the result using 1 GPU. Using 4 GPUs, the all-reduce communication accounts for 7.8\%, while the overall time reduces 69.8\% compared with the result using 2 GPUs.}
    \label{fig:time_proportion}
\end{figure}

Solving BA problems on a single GPU can suffer from computation and memory bottlenecks. We are interested in if the computation bottleneck can be resolved by adding more GPUs. We used nvprof to profile MegBA and calculated its GPU utilisation. In particular, we plot the time proportion of different computation in \sys. We repeat the same experiment with 1, 2, 4 GPUs and report the results in Figure~\ref{fig:time_proportion}. According to this figure, the profiling results show that using multiple GPUs incurs only marginal communication cost and the computation time decreases around 50\%.

\input{figures}

%% file: experiments.tex
    
\section{Experimental Results}

We evaluate the performance of \sys with the FP64, 1-GPU configuration and compare \sys against Ceres~\cite{ceres}, g2o~\cite{g2o}, RootBA~\cite{RootBA} and DeepLM~\cite{DeepLM}. \sys uses the Levenberg–Marquardt algorithm and the trust-region strategy (same as Ceres). 

\subsection{BAL Datasets}

We summarise the results of the BAL dataset in Table~\ref{tab:trafalgar} (Trafalgar), Table~\ref{tab:ladybug} (Ladybug), Table~\ref{tab:dubronvnik} (Dubrovnik), Table~\ref{tab:venice} (Venice), and Table~\ref{tab:final} (Final). As we can see in these tables, \sys consistently outperforms all baselines by up to 24x. We also evaluated PBA~\cite{pba} on the BAL dataset. PBA supports FP32 only, so we evaluate MegBA(FP32) with PBA in Table~\ref{tab:PBA}, showing that MegBA outperforms PBA by a large margin.

\input{table/Trafalgar}

\input{table/Ladybug}

\input{table/Dubrovnik}

\input{table/Venice}

\input{table/Final}

\input{table/PBA}

\subsection{1DSfM Dataset}

We further compare \sys with the baselines using the 1DSfM~\cite{1dsfm} dataset. The statistics, such as the number of observations, of the 1DSfM dataset can be found in Table~\ref{tab:1dsfm_statistics}. Results are shown in Table~\ref{tab:1dsfm}. 1DSfM is a challenging dataset and other algorithms can converge to local optima that have several magnitudes of orders larger MSE than other algorithms, while MegBA always converges to a competitive optima.

\input{table/1dsfm}
\input{table/1dsfm_mem}
\subsection{Large Synthesised Dataset}

In the end, we evaluate \sys on a larger synthetic dataset: \texttt{SynthesisedData-20000}. This dataset emulates a real-world BA problem we have in production. We use $\mathcal{U}(a, b)$ to denote a uniform distribution with a minimum as $a$ and a maximum as $b$. We generate 20,000 cameras on a circle of radius 8 uniformly. We add a uniform noise $\epsilon \sim \mathcal{U}(0, 0.01)$ to camera rotation and translation, as well as a uniform noise $\epsilon \sim \mathcal{U}(0, 0.5)$ to the camera intrinsic. We generate 80, 000 points. For each point, the $x, y$ coordinates are sampled from a uniform distribution $\mathcal{U}(-0.1, 0.1)$ and the $z$ coordinate is sampled from a uniform distribution $\mathcal{U}(-0.03, 0.03)$. The $x, y$ coordinates of each point are also added noise $\epsilon \sim \mathcal{U}(-0.1, 0.1)$. Each point can be captured by 1,000 cameras, and there are 80,000,000 observations in total.

The evaluation was conducted on a GPU server with 300 GB CPU memory in total and 32 GB GPU memory for each GPU. RootBA~\cite{RootBA} and DeepLM~\cite{DeepLM} incur out-of-memory (OOM) errors and fail to complete in this experiment. g2o cannot return results in a reasonable time (1 hour) and throws an out-of-time error.

\begin{figure}[!t]
\includegraphics[width=0.6\textwidth]{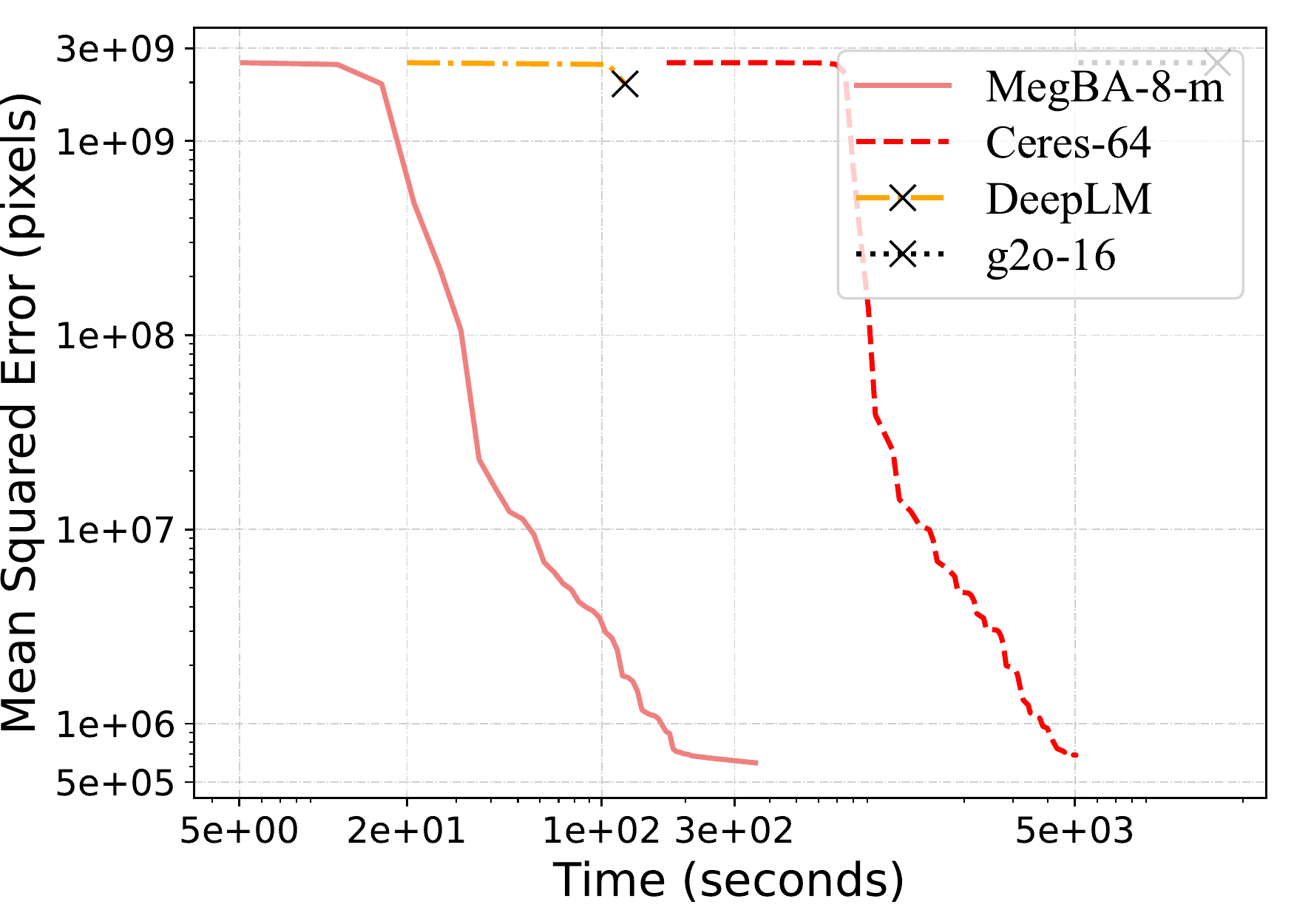}
\centering
\caption{Experimental results of the synthesised dataset. DeepLM throws an OOM error after the 6th step. g2o is significantly slower than all other baselines and we terminate its execution after 6 steps.}
\label{fig:syn}
\end{figure}

Figure~\ref{fig:syn} shows the experimental results of \sys, Ceres, DeepLM and g2o. DeepLM and g2o both cannot solve this problem due to memory and computation limitations. Only \sys and Ceres can process such a large dataset; however, \sys is 20$\times$ faster than Ceres, making \sys the state-of-the-art BA library for large BA problems (with more than 100s millions of observations). 

%% file: table/Trafalgar.tex
\begin{table}[!t]
\centering
\begin{tabular}{|l|ll|ll|ll|ll|ll|ll|}
\hline
              & \multicolumn{2}{c|}{Ceres-16}                                  & \multicolumn{2}{c|}{DeepLM}                                    & \multicolumn{2}{c|}{g2o-16}                                    & \multicolumn{2}{c|}{RootBA-16}                                 & \multicolumn{2}{c|}{MegBA-1-a}                        & \multicolumn{2}{c|}{MegBA-1-m}                                 \\ \hline
              & \multicolumn{1}{c|}{MSE}           & \multicolumn{1}{c|}{Time} & \multicolumn{1}{c|}{MSE}           & \multicolumn{1}{c|}{Time} & \multicolumn{1}{c|}{MSE}           & \multicolumn{1}{c|}{Time} & \multicolumn{1}{c|}{MSE}           & \multicolumn{1}{c|}{Time} & \multicolumn{1}{c|}{MSE}  & \multicolumn{1}{c|}{Time} & \multicolumn{1}{c|}{MSE}           & \multicolumn{1}{c|}{Time} \\ \hline
Trafalgar-21  & \multicolumn{1}{l|}{0.83}          & 0.42                      & \multicolumn{1}{l|}{0.83}          & 1.72                      & \multicolumn{1}{l|}{0.83}          & 0.79                      & \multicolumn{1}{l|}{0.83}          & \textbf{0.16}             & \multicolumn{1}{l|}{0.83} & 0.58                      & \multicolumn{1}{l|}{\textbf{0.83}} & 0.50                      \\ \hline
Trafalgar-39  & \multicolumn{1}{l|}{0.95}          & 1.02                      & \multicolumn{1}{l|}{0.95}          & 2.63                      & \multicolumn{1}{l|}{0.95}          & 2.52                      & \multicolumn{1}{l|}{0.95}          & \textbf{0.38}             & \multicolumn{1}{l|}{0.95} & 0.88                      & \multicolumn{1}{l|}{\textbf{0.95}} & 0.81                      \\ \hline
Trafalgar-50  & \multicolumn{1}{l|}{0.70}          & 1.04                      & \multicolumn{1}{l|}{0.70}          & 2.62                      & \multicolumn{1}{l|}{0.70}          & 1.90                      & \multicolumn{1}{l|}{\textbf{0.70}} & \textbf{0.44}             & \multicolumn{1}{l|}{0.70} & 0.97                      & \multicolumn{1}{l|}{\textbf{0.70}} & 0.81                      \\ \hline
Trafalgar-126 & \multicolumn{1}{l|}{0.62}          & 3.36                      & \multicolumn{1}{l|}{0.62}          & 3.98                      & \multicolumn{1}{l|}{0.62}          & 8.76                      & \multicolumn{1}{l|}{0.62}          & 8.06                      & \multicolumn{1}{l|}{0.62} & 1.83                      & \multicolumn{1}{l|}{\textbf{0.62}} & \textbf{1.52}             \\ \hline
Trafalgar-138 & \multicolumn{1}{l|}{0.53}          & 8.11                      & \multicolumn{1}{l|}{0.53}          & 4.00                      & \multicolumn{1}{l|}{\textbf{0.52}} & 12.90                     & \multicolumn{1}{l|}{0.52}          & 11.88                     & \multicolumn{1}{l|}{0.53} & 1.18                      & \multicolumn{1}{l|}{0.53}          & \textbf{1.03}             \\ \hline
Trafalgar-161 & \multicolumn{1}{l|}{0.47}          & 4.76                      & \multicolumn{1}{l|}{0.47}          & 4.11                      & \multicolumn{1}{l|}{\textbf{0.46}} & 9.04                      & \multicolumn{1}{l|}{0.46}          & 14.64                     & \multicolumn{1}{l|}{0.47} & 1.12                      & \multicolumn{1}{l|}{0.47}          & \textbf{1.11}             \\ \hline
Trafalgar-170 & \multicolumn{1}{l|}{0.47}          & 6.77                      & \multicolumn{1}{l|}{0.47}          & 4.00                      & \multicolumn{1}{l|}{\textbf{0.46}} & 14.23                     & \multicolumn{1}{l|}{0.46}          & 20.65                     & \multicolumn{1}{l|}{0.47} & 1.40                      & \multicolumn{1}{l|}{0.47}          & \textbf{1.14}             \\ \hline
Trafalgar-174 & \multicolumn{1}{l|}{0.47}          & 5.63                      & \multicolumn{1}{l|}{0.46}          & 4.17                      & \multicolumn{1}{l|}{\textbf{0.45}} & 23.42                     & \multicolumn{1}{l|}{0.46}          & 19.35                     & \multicolumn{1}{l|}{0.46} & 1.22                      & \multicolumn{1}{l|}{0.46}          & \textbf{1.19}             \\ \hline
Trafalgar-193 & \multicolumn{1}{l|}{0.46}          & 7.13                      & \multicolumn{1}{l|}{0.46}          & 4.17                      & \multicolumn{1}{l|}{\textbf{0.45}} & 16.92                     & \multicolumn{1}{l|}{0.46}          & 18.45                     & \multicolumn{1}{l|}{0.46} & 2.38                      & \multicolumn{1}{l|}{0.46}          & \textbf{2.02}             \\ \hline
Trafalgar-201 & \multicolumn{1}{l|}{0.48}          & 2.83                      & \multicolumn{1}{l|}{\textbf{0.46}} & 3.72                      & \multicolumn{1}{l|}{0.50}          & 27.17                     & \multicolumn{1}{l|}{0.46}          & 5.18                      & \multicolumn{1}{l|}{0.47} & 1.30                      & \multicolumn{1}{l|}{0.47}          & \textbf{1.22}             \\ \hline
Trafalgar-206 & \multicolumn{1}{l|}{\textbf{0.45}} & 12.80                     & \multicolumn{1}{l|}{0.45}          & 4.36                      & \multicolumn{1}{l|}{0.45}          & 22.26                     & \multicolumn{1}{l|}{0.45}          & 17.30                     & \multicolumn{1}{l|}{0.46} & 1.52                      & \multicolumn{1}{l|}{0.46}          & \textbf{1.32}             \\ \hline
Trafalgar-215 & \multicolumn{1}{l|}{0.45}          & 9.97                      & \multicolumn{1}{l|}{0.45}          & 4.33                      & \multicolumn{1}{l|}{0.45}          & 24.01                     & \multicolumn{1}{l|}{0.45}          & 12.63                     & \multicolumn{1}{l|}{0.45} & 1.24                      & \multicolumn{1}{l|}{\textbf{0.45}} & \textbf{1.21}             \\ \hline
Trafalgar-225 & \multicolumn{1}{l|}{0.44}          & 5.55                      & \multicolumn{1}{l|}{0.44}          & 4.49                      & \multicolumn{1}{l|}{0.44}          & 19.43                     & \multicolumn{1}{l|}{0.44}          & 23.96                     & \multicolumn{1}{l|}{0.44} & 1.35                      & \multicolumn{1}{l|}{\textbf{0.44}} & \textbf{1.29}             \\ \hline
Trafalgar-257 & \multicolumn{1}{l|}{0.43}          & 8.16                      & \multicolumn{1}{l|}{0.43}          & 3.82                      & \multicolumn{1}{l|}{0.43}          & 21.69                     & \multicolumn{1}{l|}{\textbf{0.43}} & 3.307                     & \multicolumn{1}{l|}{0.44} & 1.36                      & \multicolumn{1}{l|}{0.44}          & \textbf{1.15}             \\ \hline
\end{tabular}
\caption{The results of the Trafalgar dataset.}
\label{tab:trafalgar}
\end{table}

%% file: table/Ladybug.tex
\begin{table}[!t]
\centering
\begin{tabular}{|l|ll|ll|ll|ll|ll|ll|}
\hline
             & \multicolumn{2}{c|}{Ceres-16}                         & \multicolumn{2}{c|}{DeepLM}                           & \multicolumn{2}{c|}{g2o-16}                                    & \multicolumn{2}{c|}{Rootba-16}                                 & \multicolumn{2}{c|}{MegBA-1-a}                        & \multicolumn{2}{c|}{MegBA-1-m}                                 \\ \hline
             & \multicolumn{1}{c|}{MSE}  & \multicolumn{1}{c|}{Time} & \multicolumn{1}{c|}{MSE}  & \multicolumn{1}{c|}{Time} & \multicolumn{1}{c|}{MSE}           & \multicolumn{1}{c|}{Time} & \multicolumn{1}{c|}{MSE}           & \multicolumn{1}{c|}{Time} & \multicolumn{1}{c|}{MSE}  & \multicolumn{1}{c|}{Time} & \multicolumn{1}{c|}{MSE}           & \multicolumn{1}{c|}{Time} \\ \hline
Ladybug-49   & \multicolumn{1}{l|}{0.42} & 1.07                      & \multicolumn{1}{l|}{0.42} & 2.23                      & \multicolumn{1}{l|}{0.42}          & 2.50                      & \multicolumn{1}{l|}{0.42}          & 0.32                      & \multicolumn{1}{l|}{0.42} & 0.19                      & \multicolumn{1}{l|}{\textbf{0.42}} & \textbf{0.19}             \\ \hline
Ladybug-73   & \multicolumn{1}{l|}{0.37} & 1.24                      & \multicolumn{1}{l|}{0.37} & 2.72                      & \multicolumn{1}{l|}{0.37}          & 2.12                      & \multicolumn{1}{l|}{0.37}          & 0.67                      & \multicolumn{1}{l|}{0.37} & 0.24                      & \multicolumn{1}{l|}{\textbf{0.37}} & \textbf{0.22}             \\ \hline
Ladybug-138  & \multicolumn{1}{l|}{0.70} & 2.72                      & \multicolumn{1}{l|}{0.70} & 3.00                      & \multicolumn{1}{l|}{0.70}          & 31.21                     & \multicolumn{1}{l|}{0.82}          & 0.79                      & \multicolumn{1}{l|}{0.70} & 0.32                      & \multicolumn{1}{l|}{\textbf{0.70}} & \textbf{0.29}             \\ \hline
Ladybug-318  & \multicolumn{1}{l|}{0.48} & 3.04                      & \multicolumn{1}{l|}{0.48} & 2.95                      & \multicolumn{1}{l|}{0.48}          & 65.70                     & \multicolumn{1}{l|}{0.48}          & 3.78                      & \multicolumn{1}{l|}{0.48} & 0.44                      & \multicolumn{1}{l|}{\textbf{0.48}} & \textbf{0.33}             \\ \hline
Ladybug-372  & \multicolumn{1}{l|}{0.55} & 4.45                      & \multicolumn{1}{l|}{0.55} & 2.10                      & \multicolumn{1}{l|}{0.55}          & 83.52                     & \multicolumn{1}{l|}{0.55}          & 2.96                      & \multicolumn{1}{l|}{0.55} & 0.54                      & \multicolumn{1}{l|}{\textbf{0.55}} & \textbf{0.41}             \\ \hline
Ladybug-412  & \multicolumn{1}{l|}{0.50} & 4.07                      & \multicolumn{1}{l|}{0.50} & 3.34                      & \multicolumn{1}{l|}{0.49}          & 86.94                     & \multicolumn{1}{l|}{0.50}          & 2.83                      & \multicolumn{1}{l|}{0.49} & 0.58                      & \multicolumn{1}{l|}{\textbf{0.49}} & \textbf{0.44}             \\ \hline
Ladybug-460  & \multicolumn{1}{l|}{0.53} & 6.02                      & \multicolumn{1}{l|}{0.53} & 3.47                      & \multicolumn{1}{l|}{0.53}          & 243.32                    & \multicolumn{1}{l|}{0.53}          & 5.38                      & \multicolumn{1}{l|}{0.53} & 0.69                      & \multicolumn{1}{l|}{\textbf{0.53}} & \textbf{0.55}             \\ \hline
Ladybug-539  & \multicolumn{1}{l|}{0.55} & 8.37                      & \multicolumn{1}{l|}{0.55} & 3.85                      & \multicolumn{1}{l|}{0.55}          & 65.63                     & \multicolumn{1}{l|}{0.55}          & 5.08                      & \multicolumn{1}{l|}{0.55} & 0.75                      & \multicolumn{1}{l|}{\textbf{0.55}} & \textbf{0.57}             \\ \hline
Ladybug-598  & \multicolumn{1}{l|}{0.55} & 8.54                      & \multicolumn{1}{l|}{0.55} & 3.34                      & \multicolumn{1}{l|}{\textbf{0.54}} & 293.69                    & \multicolumn{1}{l|}{0.55}          & 4.16                      & \multicolumn{1}{l|}{0.55} & 0.79                      & \multicolumn{1}{l|}{0.55}          & \textbf{0.60}             \\ \hline
Ladybug-646  & \multicolumn{1}{l|}{0.55} & 9.21                      & \multicolumn{1}{l|}{0.55} & 3.56                      & \multicolumn{1}{l|}{0.56}          & 200.83                    & \multicolumn{1}{l|}{0.55}          & 5.44                      & \multicolumn{1}{l|}{0.55} & 0.92                      & \multicolumn{1}{l|}{\textbf{0.55}} & \textbf{0.74}             \\ \hline
Ladybug-707  & \multicolumn{1}{l|}{0.57} & 15.30                     & \multicolumn{1}{l|}{0.56} & 4.48                      & \multicolumn{1}{l|}{0.56}          & 283.15                    & \multicolumn{1}{l|}{0.56}          & 6.59                      & \multicolumn{1}{l|}{0.56} & 0.98                      & \multicolumn{1}{l|}{\textbf{0.56}} & \textbf{0.76}             \\ \hline
Ladybug-783  & \multicolumn{1}{l|}{0.53} & 7.69                      & \multicolumn{1}{l|}{0.53} & 5.01                      & \multicolumn{1}{l|}{0.53}          & 539.85                    & \multicolumn{1}{l|}{\textbf{0.52}} & 15.95                     & \multicolumn{1}{l|}{0.53} & 0.98                      & \multicolumn{1}{l|}{0.53}          & \textbf{0.78}             \\ \hline
Ladybug-810  & \multicolumn{1}{l|}{0.52} & 9.67                      & \multicolumn{1}{l|}{0.52} & 4.77                      & \multicolumn{1}{l|}{0.52}          & 391.78                    & \multicolumn{1}{l|}{0.52}          & 13.71                     & \multicolumn{1}{l|}{0.52} & 1.00                      & \multicolumn{1}{l|}{\textbf{0.52}} & \textbf{0.74}             \\ \hline
Ladybug-856  & \multicolumn{1}{l|}{0.51} & 13.30                     & \multicolumn{1}{l|}{0.51} & 4.81                      & \multicolumn{1}{l|}{0.51}          & 114.89                    & \multicolumn{1}{l|}{0.51}          & 8.01                      & \multicolumn{1}{l|}{0.51} & 1.10                      & \multicolumn{1}{l|}{\textbf{0.51}} & \textbf{0.86}             \\ \hline
Ladybug-885  & \multicolumn{1}{l|}{0.51} & 15.00                     & \multicolumn{1}{l|}{0.51} & 5.08                      & \multicolumn{1}{l|}{0.51}          & 308.21                    & \multicolumn{1}{l|}{0.51}          & 8.03                      & \multicolumn{1}{l|}{0.51} & 1.01                      & \multicolumn{1}{l|}{\textbf{0.51}} & \textbf{0.79}             \\ \hline
Ladybug-931  & \multicolumn{1}{l|}{0.55} & 15.30                     & \multicolumn{1}{l|}{0.55} & 4.56                      & \multicolumn{1}{l|}{0.55}          & 220.54                    & \multicolumn{1}{l|}{0.55}          & 11.08                     & \multicolumn{1}{l|}{0.55} & 1.18                      & \multicolumn{1}{l|}{\textbf{0.55}} & \textbf{0.96}             \\ \hline
Ladybug-969  & \multicolumn{1}{l|}{0.55} & 17.50                     & \multicolumn{1}{l|}{0.54} & 5.21                      & \multicolumn{1}{l|}{0.55}          & 197.59                    & \multicolumn{1}{l|}{0.54}          & 7.79                      & \multicolumn{1}{l|}{0.54} & 1.19                      & \multicolumn{1}{l|}{\textbf{0.54}} & \textbf{0.94}             \\ \hline
Ladybug-1031 & \multicolumn{1}{l|}{0.55} & 12.90                     & \multicolumn{1}{l|}{0.55} & 4.04                      & \multicolumn{1}{l|}{0.55}          & 203.63                    & \multicolumn{1}{l|}{0.55}          & 10.24                     & \multicolumn{1}{l|}{0.55} & 1.48                      & \multicolumn{1}{l|}{\textbf{0.55}} & \textbf{1.18}             \\ \hline
Ladybug-1064 & \multicolumn{1}{l|}{0.55} & 15.40                     & \multicolumn{1}{l|}{0.56} & 3.17                      & \multicolumn{1}{l|}{0.55}          & 355.78                    & \multicolumn{1}{l|}{0.55}          & 4.47                      & \multicolumn{1}{l|}{0.55} & 1.23                      & \multicolumn{1}{l|}{\textbf{0.55}} & \textbf{1.03}             \\ \hline
Ladybug-1118 & \multicolumn{1}{l|}{0.57} & 15.40                     & \multicolumn{1}{l|}{0.58} & 3.68                      & \multicolumn{1}{l|}{0.58}          & 311.09                    & \multicolumn{1}{l|}{0.58}          & 6.16                      & \multicolumn{1}{l|}{0.57} & 1.39                      & \multicolumn{1}{l|}{\textbf{0.57}} & \textbf{1.11}             \\ \hline
Ladybug-1152 & \multicolumn{1}{l|}{0.56} & 13.70                     & \multicolumn{1}{l|}{0.56} & 3.07                      & \multicolumn{1}{l|}{0.56}          & 488.39                    & \multicolumn{1}{l|}{0.56}          & 7.53                      & \multicolumn{1}{l|}{0.56} & 1.46                      & \multicolumn{1}{l|}{\textbf{0.56}} & \textbf{1.10}             \\ \hline
Ladybug-1197 & \multicolumn{1}{l|}{0.57} & 17.30                     & \multicolumn{1}{l|}{0.57} & 4.13                      & \multicolumn{1}{l|}{0.57}          & 122.18                    & \multicolumn{1}{l|}{0.57}          & 8.58                      & \multicolumn{1}{l|}{0.57} & 1.40                      & \multicolumn{1}{l|}{\textbf{0.57}} & \textbf{1.13}             \\ \hline
Ladybug-1235 & \multicolumn{1}{l|}{0.56} & 20.30                     & \multicolumn{1}{l|}{0.58} & 3.34                      & \multicolumn{1}{l|}{0.56}          & 212.62                    & \multicolumn{1}{l|}{0.57}          & 5.89                      & \multicolumn{1}{l|}{0.56} & 1.30                      & \multicolumn{1}{l|}{\textbf{0.56}} & \textbf{1.06}             \\ \hline
Ladybug-1266 & \multicolumn{1}{l|}{0.57} & 23.60                     & \multicolumn{1}{l|}{0.56} & 4.53                      & \multicolumn{1}{l|}{0.56}          & 342.70                    & \multicolumn{1}{l|}{0.56}          & 8.81                      & \multicolumn{1}{l|}{0.56} & 1.44                      & \multicolumn{1}{l|}{\textbf{0.56}} & \textbf{1.18}             \\ \hline
Ladybug-1340 & \multicolumn{1}{l|}{0.57} & 26.20                     & \multicolumn{1}{l|}{0.57} & 4.54                      & \multicolumn{1}{l|}{2.02}          & 453.25                    & \multicolumn{1}{l|}{0.57}          & 9.17                      & \multicolumn{1}{l|}{0.57} & 1.61                      & \multicolumn{1}{l|}{\textbf{0.57}} & \textbf{1.31}             \\ \hline
Ladybug-1469 & \multicolumn{1}{l|}{0.56} & 26.30                     & \multicolumn{1}{l|}{0.57} & 3.84                      & \multicolumn{1}{l|}{0.58}          & 724.13                    & \multicolumn{1}{l|}{0.56}          & 8.30                      & \multicolumn{1}{l|}{0.56} & 1.73                      & \multicolumn{1}{l|}{\textbf{0.56}} & \textbf{1.37}             \\ \hline
Ladybug-1514 & \multicolumn{1}{l|}{0.56} & 25.20                     & \multicolumn{1}{l|}{0.56} & 4.69                      & \multicolumn{1}{l|}{1.18}          & 459.05                    & \multicolumn{1}{l|}{0.56}          & 8.61                      & \multicolumn{1}{l|}{0.56} & 1.59                      & \multicolumn{1}{l|}{\textbf{0.56}} & \textbf{1.34}             \\ \hline
Ladybug-1587 & \multicolumn{1}{l|}{0.59} & 37.50                     & \multicolumn{1}{l|}{0.56} & 4.82                      & \multicolumn{1}{l|}{1.15}          & 567.71                    & \multicolumn{1}{l|}{0.58}          & 6.77                      & \multicolumn{1}{l|}{0.56} & 2.01                      & \multicolumn{1}{l|}{\textbf{0.56}} & \textbf{1.58}             \\ \hline
Ladybug-1642 & \multicolumn{1}{l|}{0.58} & 36.10                     & \multicolumn{1}{l|}{0.56} & 3.64                      & \multicolumn{1}{l|}{0.76}          & 127.00                    & \multicolumn{1}{l|}{0.56}          & 16.12                     & \multicolumn{1}{l|}{0.56} & 1.73                      & \multicolumn{1}{l|}{\textbf{0.56}} & \textbf{1.43}             \\ \hline
Ladybug-1695 & \multicolumn{1}{l|}{0.56} & 32.30                     & \multicolumn{1}{l|}{0.56} & 4.24                      & \multicolumn{1}{l|}{0.62}          & 799.46                    & \multicolumn{1}{l|}{0.59}          & 5.46                      & \multicolumn{1}{l|}{0.56} & 1.87                      & \multicolumn{1}{l|}{\textbf{0.56}} & \textbf{1.51}             \\ \hline
Ladybug-1723 & \multicolumn{1}{l|}{0.56} & 34.50                     & \multicolumn{1}{l|}{0.57} & 3.93                      & \multicolumn{1}{l|}{1.96}          & 140.69                    & \multicolumn{1}{l|}{0.56}          & 7.05                      & \multicolumn{1}{l|}{0.56} & 0.93                      & \multicolumn{1}{l|}{\textbf{0.56}} & \textbf{0.77}             \\ \hline
\end{tabular}
\caption{The results of the Ladybug dataset.}
\label{tab:ladybug}
\end{table}

%% file: table/Dubrovnik.tex
\begin{table}[!t]
\centering
\begin{tabular}{|l|ll|ll|ll|ll|ll|ll|}
\hline
              & \multicolumn{2}{c|}{Ceres-16}                                  & \multicolumn{2}{c|}{DeepLM}                           & \multicolumn{2}{c|}{g2o-16}                           & \multicolumn{2}{c|}{Rootba-16}                        & \multicolumn{2}{c|}{MegBA-1-a}                        & \multicolumn{2}{c|}{MegBA-1-m}                                 \\ \hline
              & \multicolumn{1}{c|}{MSE}           & \multicolumn{1}{c|}{Time} & \multicolumn{1}{c|}{MSE}  & \multicolumn{1}{c|}{Time} & \multicolumn{1}{c|}{MSE}  & \multicolumn{1}{c|}{Time} & \multicolumn{1}{c|}{MSE}  & \multicolumn{1}{c|}{Time} & \multicolumn{1}{c|}{MSE}  & \multicolumn{1}{c|}{Time} & \multicolumn{1}{c|}{MSE}           & \multicolumn{1}{c|}{Time} \\ \hline
Dubrovnik-16  & \multicolumn{1}{l|}{0.22}          & 0.79                      & \multicolumn{1}{l|}{0.22} & 1.75                      & \multicolumn{1}{l|}{0.22} & 1.72                      & \multicolumn{1}{l|}{0.22} & \textbf{0.36}                      & \multicolumn{1}{l|}{0.22} & 1.32                      & \multicolumn{1}{l|}{\textbf{0.22}} & 1.24             \\ \hline
Dubrovnik-88  & \multicolumn{1}{l|}{0.75}          & 7.69                      & \multicolumn{1}{l|}{0.75} & 4.77                      & \multicolumn{1}{l|}{0.75} & 21.87                     & \multicolumn{1}{l|}{0.75} & 11.14                     & \multicolumn{1}{l|}{0.75} & 1.89                      & \multicolumn{1}{l|}{\textbf{0.75}} & \textbf{1.63}             \\ \hline
Dubrovnik-135 & \multicolumn{1}{l|}{0.67}          & 17.00                     & \multicolumn{1}{l|}{0.67} & 5.88                      & \multicolumn{1}{l|}{0.67} & 34.21                     & \multicolumn{1}{l|}{0.67} & 17.99                     & \multicolumn{1}{l|}{0.67} & 1.52                      & \multicolumn{1}{l|}{\textbf{0.67}} & \textbf{1.50}             \\ \hline
Dubrovnik-142 & \multicolumn{1}{l|}{0.48}          & 17.30                     & \multicolumn{1}{l|}{0.48} & 6.03                      & \multicolumn{1}{l|}{0.48} & 30.11                     & \multicolumn{1}{l|}{0.48} & 6.74                      & \multicolumn{1}{l|}{0.48} & 2.34                      & \multicolumn{1}{l|}{\textbf{0.48}} & \textbf{2.25}             \\ \hline
Dubrovnik-150 & \multicolumn{1}{l|}{0.43}          & 15.20                     & \multicolumn{1}{l|}{0.43} & 6.27                      & \multicolumn{1}{l|}{0.43} & 31.32                     & \multicolumn{1}{l|}{0.43} & 33.01                     & \multicolumn{1}{l|}{0.43} & 1.11                      & \multicolumn{1}{l|}{\textbf{0.43}} & \textbf{1.00}             \\ \hline
Dubrovnik-161 & \multicolumn{1}{l|}{0.41}          & 16.80                     & \multicolumn{1}{l|}{0.41} & 7.61                      & \multicolumn{1}{l|}{0.41} & 31.96                     & \multicolumn{1}{l|}{0.41} & 19.32                     & \multicolumn{1}{l|}{0.41} & 0.71                      & \multicolumn{1}{l|}{\textbf{0.41}} & \textbf{0.69}             \\ \hline
Dubrovnik-173 & \multicolumn{1}{l|}{0.41}          & 15.40                     & \multicolumn{1}{l|}{0.41} & 7.38                      & \multicolumn{1}{l|}{0.41} & 38.44                     & \multicolumn{1}{l|}{0.41} & 59.45                     & \multicolumn{1}{l|}{0.41} & 0.96                      & \multicolumn{1}{l|}{\textbf{0.41}} & \textbf{0.84}             \\ \hline
Dubrovnik-182 & \multicolumn{1}{l|}{0.45}          & 13.40                     & \multicolumn{1}{l|}{0.45} & 8.11                      & \multicolumn{1}{l|}{0.45} & 41.91                     & \multicolumn{1}{l|}{0.45} & 67.52                     & \multicolumn{1}{l|}{0.45} & 1.08                      & \multicolumn{1}{l|}{\textbf{0.45}} & \textbf{1.00}             \\ \hline
Dubrovnik-202 & \multicolumn{1}{l|}{0.43}          & 21.70                     & \multicolumn{1}{l|}{0.43} & 9.40                      & \multicolumn{1}{l|}{0.44} & 54.12                     & \multicolumn{1}{l|}{0.43} & 22.19                     & \multicolumn{1}{l|}{0.43} & 1.41                      & \multicolumn{1}{l|}{\textbf{0.43}} & \textbf{1.29}             \\ \hline
Dubrovnik-237 & \multicolumn{1}{l|}{0.42}          & 25.50                     & \multicolumn{1}{l|}{0.42} & 9.95                      & \multicolumn{1}{l|}{0.42} & 59.35                     & \multicolumn{1}{l|}{0.42} & 17.27                     & \multicolumn{1}{l|}{0.41} & 0.91                      & \multicolumn{1}{l|}{\textbf{0.41}} & \textbf{0.80}             \\ \hline
Dubrovnik-253 & \multicolumn{1}{l|}{0.38}          & 29.00                     & \multicolumn{1}{l|}{0.38} & 10.82                     & \multicolumn{1}{l|}{0.39} & 69.64                     & \multicolumn{1}{l|}{0.38} & 50.39                     & \multicolumn{1}{l|}{0.38} & 1.45                      & \multicolumn{1}{l|}{\textbf{0.38}} & \textbf{1.24}             \\ \hline
Dubrovnik-262 & \multicolumn{1}{l|}{0.37}          & 95.90                     & \multicolumn{1}{l|}{0.37} & 11.25                     & \multicolumn{1}{l|}{0.37} & 64.68                     & \multicolumn{1}{l|}{0.37} & 52.60                     & \multicolumn{1}{l|}{0.37} & 2.79                      & \multicolumn{1}{l|}{\textbf{0.37}} & \textbf{2.59}             \\ \hline
Dubrovnik-273 & \multicolumn{1}{l|}{0.37}          & 47.90                     & \multicolumn{1}{l|}{0.37} & 11.41                     & \multicolumn{1}{l|}{0.37} & 63.24                     & \multicolumn{1}{l|}{0.36} & 60.37                     & \multicolumn{1}{l|}{0.37} & 1.87                      & \multicolumn{1}{l|}{\textbf{0.37}} & \textbf{1.78}             \\ \hline
Dubrovnik-287 & \multicolumn{1}{l|}{0.36}          & 26.80                     & \multicolumn{1}{l|}{0.36} & 11.43                     & \multicolumn{1}{l|}{0.37} & 93.30                     & \multicolumn{1}{l|}{0.36} & 40.84                     & \multicolumn{1}{l|}{0.36} & 2.35                      & \multicolumn{1}{l|}{\textbf{0.36}} & \textbf{2.13}             \\ \hline
Dubrovnik-308 & \multicolumn{1}{l|}{0.37}          & 33.80                     & \multicolumn{1}{l|}{0.37} & 11.22                     & \multicolumn{1}{l|}{0.37} & 91.82                     & \multicolumn{1}{l|}{0.37} & 37.15                     & \multicolumn{1}{l|}{0.37} & 3.23                      & \multicolumn{1}{l|}{\textbf{0.37}} & \textbf{2.88}             \\ \hline
Dubrovnik-356 & \multicolumn{1}{l|}{\textbf{0.39}} & 116.00                    & \multicolumn{1}{l|}{0.40} & 6.12                      & \multicolumn{1}{l|}{0.39} & 94.39                     & \multicolumn{1}{l|}{0.39} & 78.16                     & \multicolumn{1}{l|}{0.41} & 3.64                      & \multicolumn{1}{l|}{0.41}          & \textbf{3.26}             \\ \hline
\end{tabular}
\caption{The results of the Dubrovnik dataset.}
\label{tab:dubronvnik}
\end{table}

%% file: table/Venice.tex
\begin{table}[!t]
\centering
\begin{tabular}{|l|ll|ll|ll|ll|ll|ll|}
\hline
            & \multicolumn{2}{c|}{Ceres-16}                                  & \multicolumn{2}{c|}{DeepLM}                           & \multicolumn{2}{c|}{g2o-16}                                    & \multicolumn{2}{c|}{Rootba-16}                                 & \multicolumn{2}{c|}{MegBA-1-a}                        & \multicolumn{2}{c|}{MegBA-1-m}                                 \\ \hline
            & \multicolumn{1}{c|}{MSE}           & \multicolumn{1}{c|}{Time} & \multicolumn{1}{c|}{MSE}  & \multicolumn{1}{c|}{Time} & \multicolumn{1}{c|}{MSE}           & \multicolumn{1}{c|}{Time} & \multicolumn{1}{c|}{MSE}           & \multicolumn{1}{c|}{Time} & \multicolumn{1}{c|}{MSE}  & \multicolumn{1}{c|}{Time} & \multicolumn{1}{c|}{MSE}           & \multicolumn{1}{c|}{Time} \\ \hline
Venice-52   & \multicolumn{1}{l|}{0.75}          & 7.95                      & \multicolumn{1}{l|}{0.75} & 4.66                      & \multicolumn{1}{l|}{\textbf{0.68}} & 17.94                     & \multicolumn{1}{l|}{0.68}          & 2.39                      & \multicolumn{1}{l|}{0.73} & 1.85                      & \multicolumn{1}{l|}{0.73}          & \textbf{1.59}             \\ \hline
Venice-89   & \multicolumn{1}{l|}{0.50}          & 8.47                      & \multicolumn{1}{l|}{0.50} & 6.41                      & \multicolumn{1}{l|}{0.50}          & 27.53                     & \multicolumn{1}{l|}{0.51}          & \textbf{4.51}                      & \multicolumn{1}{l|}{0.50} & 5.05                      & \multicolumn{1}{l|}{\textbf{0.50}} & 4.85             \\ \hline
Venice-245  & \multicolumn{1}{l|}{0.84}          & 31.90                     & \multicolumn{1}{l|}{0.87} & 5.66                      & \multicolumn{1}{l|}{0.87}          & 123.27                    & \multicolumn{1}{l|}{0.85}          & 10.47                     & \multicolumn{1}{l|}{0.84} & 3.26                      & \multicolumn{1}{l|}{\textbf{0.84}} & \textbf{3.06}             \\ \hline
Venice-427  & \multicolumn{1}{l|}{0.63}          & 58.20                     & \multicolumn{1}{l|}{0.63} & 15.46                     & \multicolumn{1}{l|}{\textbf{0.62}} & 115.24                    & \multicolumn{1}{l|}{0.63}          & 43.12                     & \multicolumn{1}{l|}{0.63} & 5.17                      & \multicolumn{1}{l|}{0.63}          & \textbf{5.17}             \\ \hline
Venice-744  & \multicolumn{1}{l|}{0.51}          & 109.0                     & \multicolumn{1}{l|}{0.51} & 27.79                     & \multicolumn{1}{l|}{0.51}          & 659.83                    & \multicolumn{1}{l|}{0.51}          & 43.60                     & \multicolumn{1}{l|}{0.50} & 6.26                      & \multicolumn{1}{l|}{\textbf{0.50}} & \textbf{6.04}             \\ \hline
Venice-951  & \multicolumn{1}{l|}{\textbf{0.44}} & 84.30                     & \multicolumn{1}{l|}{F}    & F                         & \multicolumn{1}{l|}{0.57}          & 793.15                    & \multicolumn{1}{l|}{0.45}          & 59.45                     & \multicolumn{1}{l|}{0.45} & 13.80                     & \multicolumn{1}{l|}{0.45}          & \textbf{13.64}            \\ \hline
Venice-1102 & \multicolumn{1}{l|}{0.40}          & 146.0                     & \multicolumn{1}{l|}{0.39} & 32.80                     & \multicolumn{1}{l|}{0.56}          & 585.86                    & \multicolumn{1}{l|}{0.40}          & 94.13                     & \multicolumn{1}{l|}{0.39} & 14.42                     & \multicolumn{1}{l|}{\textbf{0.39}} & \textbf{13.24}            \\ \hline
Venice-1158 & \multicolumn{1}{l|}{0.42}          & 33.10                     & \multicolumn{1}{l|}{0.43} & 12.07                     & \multicolumn{1}{l|}{0.48}          & 581.79                    & \multicolumn{1}{l|}{0.36}          & 51.18                     & \multicolumn{1}{l|}{0.36} & 6.74                      & \multicolumn{1}{l|}{\textbf{0.36}} & \textbf{6.32}             \\ \hline
Venice-1184 & \multicolumn{1}{l|}{0.77}          & 30.10                     & \multicolumn{1}{l|}{0.35} & 36.32                     & \multicolumn{1}{l|}{0.46}          & 600.57                    & \multicolumn{1}{l|}{0.35}          & 38.16                     & \multicolumn{1}{l|}{0.34} & 36.24                     & \multicolumn{1}{l|}{\textbf{0.34}}              & \textbf{28.52}                           \\ \hline
Venice-1238 & \multicolumn{1}{l|}{0.35}          & 71.40                     & \multicolumn{1}{l|}{0.34} & 13.20                     & \multicolumn{1}{l|}{0.46}          & 531.25                    & \multicolumn{1}{l|}{0.34}          & 90.36                     & \multicolumn{1}{l|}{0.34} & 8.36                      & \multicolumn{1}{l|}{\textbf{0.34}} & \textbf{7.75}             \\ \hline
Venice-1288 & \multicolumn{1}{l|}{0.33}          & 153.0                     & \multicolumn{1}{l|}{0.33} & 19.79                     & \multicolumn{1}{l|}{0.45}          & 630.90                    & \multicolumn{1}{l|}{0.33}          & 95.45                     & \multicolumn{1}{l|}{0.33} & 8.36                      & \multicolumn{1}{l|}{\textbf{0.33}} & \textbf{7.64}             \\ \hline
Venice-1350 & \multicolumn{1}{l|}{0.34}          & 52.30                     & \multicolumn{1}{l|}{0.34} & 10.47                     & \multicolumn{1}{l|}{0.44}          & 551.41                    & \multicolumn{1}{l|}{0.34}          & 53.85                     & \multicolumn{1}{l|}{0.33} & 6.79                      & \multicolumn{1}{l|}{\textbf{0.33}} & \textbf{6.43}             \\ \hline
Venice-1408 & \multicolumn{1}{l|}{0.35}          & 76.50                     & \multicolumn{1}{l|}{0.35} & 10.72                     & \multicolumn{1}{l|}{0.47}          & 687.48                    & \multicolumn{1}{l|}{0.35}          & 47.00                     & \multicolumn{1}{l|}{0.34} & 5.66                      & \multicolumn{1}{l|}{\textbf{0.34}} & \textbf{5.34}             \\ \hline
Venice-1425 & \multicolumn{1}{l|}{0.34}          & 153.0                     & \multicolumn{1}{l|}{0.34} & 10.80                     & \multicolumn{1}{l|}{0.44}          & 635.88                    & \multicolumn{1}{l|}{0.34}          & 183.28                    & \multicolumn{1}{l|}{0.34} & 5.67                      & \multicolumn{1}{l|}{\textbf{0.34}} & \textbf{5.28}             \\ \hline
Venice-1473 & \multicolumn{1}{l|}{0.33}          & 110.0                     & \multicolumn{1}{l|}{0.33} & 43.24                     & \multicolumn{1}{l|}{0.42}          & 571.33                    & \multicolumn{1}{l|}{0.33}          & 61.21                     & \multicolumn{1}{l|}{0.33} & 7.05                      & \multicolumn{1}{l|}{\textbf{0.33}} & \textbf{6.68}             \\ \hline
Venice-1490 & \multicolumn{1}{l|}{0.33}          & 66.80                     & \multicolumn{1}{l|}{0.33} & 14.25                     & \multicolumn{1}{l|}{0.33}          & 571.17                    & \multicolumn{1}{l|}{0.33}          & 123.52                    & \multicolumn{1}{l|}{0.32} & 5.64                      & \multicolumn{1}{l|}{\textbf{0.32}} & \textbf{5.32}             \\ \hline
Venice-1521 & \multicolumn{1}{l|}{0.33}          & 75.60                     & \multicolumn{1}{l|}{0.33} & 10.98                     & \multicolumn{1}{l|}{0.33}          & 666.63                    & \multicolumn{1}{l|}{0.33}          & 93.50                     & \multicolumn{1}{l|}{0.32} & 7.47                      & \multicolumn{1}{l|}{\textbf{0.32}} & \textbf{6.78}             \\ \hline
Venice-1544 & \multicolumn{1}{l|}{0.33}          & 173.0                     & \multicolumn{1}{l|}{0.33} & 26.59                     & \multicolumn{1}{l|}{0.42}          & 653.50                    & \multicolumn{1}{l|}{0.33}          & 132.69                    & \multicolumn{1}{l|}{0.32} & 5.71                      & \multicolumn{1}{l|}{\textbf{0.32}} & \textbf{5.33}             \\ \hline
Venice-1638 & \multicolumn{1}{l|}{0.57}          & 75.40                     & \multicolumn{1}{l|}{0.58} & 30.04                     & \multicolumn{1}{l|}{\textbf{0.56}} & 1015                      & \multicolumn{1}{l|}{0.57}          & 54.58                     & \multicolumn{1}{l|}{0.58} & 17.66                     & \multicolumn{1}{l|}{0.58}          & \textbf{16.60}            \\ \hline
Venice-1666 & \multicolumn{1}{l|}{0.48}          & 74.00                     & \multicolumn{1}{l|}{0.48} & 23.48                     & \multicolumn{1}{l|}{0.45}          & 1033                      & \multicolumn{1}{l|}{\textbf{0.43}} & 85.30                     & \multicolumn{1}{l|}{0.44} & 10.75                     & \multicolumn{1}{l|}{0.44}          & \textbf{10.08}            \\ \hline
Venice-1672 & \multicolumn{1}{l|}{0.38}          & 260.0                     & \multicolumn{1}{l|}{0.38} & 33.49                     & \multicolumn{1}{l|}{0.38}          & 991.73                    & \multicolumn{1}{l|}{\textbf{0.37}} & 123.52                    & \multicolumn{1}{l|}{0.40} & 12.20                     & \multicolumn{1}{l|}{0.40}          & \textbf{11.06}            \\ \hline
Venice-1681 & \multicolumn{1}{l|}{0.37}          & 48.40                     & \multicolumn{1}{l|}{0.34} & 46.98                     & \multicolumn{1}{l|}{0.34}          & 1067                      & \multicolumn{1}{l|}{0.34}          & 118.23                    & \multicolumn{1}{l|}{0.34} & 9.16                      & \multicolumn{1}{l|}{\textbf{0.34}} & \textbf{8.38}             \\ \hline
Venice-1682 & \multicolumn{1}{l|}{0.34}          & 164.0                     & \multicolumn{1}{l|}{0.33} & 38.92                     & \multicolumn{1}{l|}{0.34}          & 956.35                    & \multicolumn{1}{l|}{0.33}          & 274.34                    & \multicolumn{1}{l|}{0.33} & 17.39                     & \multicolumn{1}{l|}{\textbf{0.33}} & \textbf{15.78}            \\ \hline
Venice-1684 & \multicolumn{1}{l|}{0.34}          & 207.0                     & \multicolumn{1}{l|}{0.33} & 43.94                     & \multicolumn{1}{l|}{0.33}          & 1658                      & \multicolumn{1}{l|}{0.33}          & 202.10                    & \multicolumn{1}{l|}{0.33} & 11.78                     & \multicolumn{1}{l|}{\textbf{0.33}} & \textbf{11.25}            \\ \hline
Venice-1695 & \multicolumn{1}{l|}{0.37}          & 53.40                     & \multicolumn{1}{l|}{0.34} & 44.99                     & \multicolumn{1}{l|}{0.34}          & 1648                      & \multicolumn{1}{l|}{\textbf{0.33}} & 292.24                    & \multicolumn{1}{l|}{0.34} & 16.16                     & \multicolumn{1}{l|}{0.34}          & \textbf{14.48}            \\ \hline
Venice-1696 & \multicolumn{1}{l|}{0.38}          & 43.00                     & \multicolumn{1}{l|}{0.33} & 47.04                     & \multicolumn{1}{l|}{0.34}          & 1335                      & \multicolumn{1}{l|}{\textbf{0.33}} & 243.68                    & \multicolumn{1}{l|}{0.34} & 7.43                      & \multicolumn{1}{l|}{0.34}          & \textbf{7.06}             \\ \hline
Venice-1706 & \multicolumn{1}{l|}{0.34}          & 234.0                     & \multicolumn{1}{l|}{0.34} & 38.17                     & \multicolumn{1}{l|}{\textbf{0.33}} & 1845                      & \multicolumn{1}{l|}{0.34}          & 196.48                    & \multicolumn{1}{l|}{0.34} & 19.72                     & \multicolumn{1}{l|}{0.34}          & \textbf{18.59}            \\ \hline
Venice-1776 & \multicolumn{1}{l|}{0.33}          & 192.0                     & \multicolumn{1}{l|}{0.33} & 46.69                     & \multicolumn{1}{l|}{0.35}          & 1147                      & \multicolumn{1}{l|}{0.34}          & 62.10                     & \multicolumn{1}{l|}{0.33} & 10.56                     & \multicolumn{1}{l|}{\textbf{0.33}} & \textbf{10.02}            \\ \hline
Venice-1778 & \multicolumn{1}{l|}{0.33}          & 319.0                     & \multicolumn{1}{l|}{0.33} & 24.44                     & \multicolumn{1}{l|}{0.34}          & 890.55                    & \multicolumn{1}{l|}{0.34}          & 73.94                     & \multicolumn{1}{l|}{0.33} & 11.96                     & \multicolumn{1}{l|}{\textbf{0.33}} & \textbf{10.92}            \\ \hline
\end{tabular}
\caption{The results of the Venice dataset. F stands for failed.}
\label{tab:venice}
\end{table}

%% file: table/Final.tex
\begin{table}[!t]
\centering
\begin{tabular}{|l|ll|ll|ll|ll|ll|ll|}
\hline
           & \multicolumn{2}{c|}{Ceres-16}                         & \multicolumn{2}{c|}{DeepLM}                                    & \multicolumn{2}{c|}{g2o-16}                           & \multicolumn{2}{c|}{Rootba-16}                        & \multicolumn{2}{c|}{MegBA-1-a}                        & \multicolumn{2}{c|}{MegBA-1-m}                                 \\ \hline
           & \multicolumn{1}{c|}{MSE}  & \multicolumn{1}{c|}{Time} & \multicolumn{1}{c|}{MSE}           & \multicolumn{1}{c|}{Time} & \multicolumn{1}{c|}{MSE}  & \multicolumn{1}{c|}{Time} & \multicolumn{1}{c|}{MSE}  & \multicolumn{1}{c|}{Time} & \multicolumn{1}{c|}{MSE}  & \multicolumn{1}{c|}{Time} & \multicolumn{1}{c|}{MSE}           & \multicolumn{1}{c|}{Time} \\ \hline
Final-93   & \multicolumn{1}{l|}{0.51} & 4.10                      & \multicolumn{1}{l|}{0.51}          & 1.56                      & \multicolumn{1}{l|}{0.51} & 6.22                      & \multicolumn{1}{l|}{0.51} & 1.84                      & \multicolumn{1}{l|}{0.51} & 0.78                      & \multicolumn{1}{l|}{\textbf{0.51}} & \textbf{0.67}             \\ \hline
Final-394  & \multicolumn{1}{l|}{0.56} & 30.10                     & \multicolumn{1}{l|}{0.56}          & 5.10                      & \multicolumn{1}{l|}{0.56} & 101.72                    & \multicolumn{1}{l|}{0.56} & 20.07                     & \multicolumn{1}{l|}{0.56} & 1.10                      & \multicolumn{1}{l|}{\textbf{0.56}} & \textbf{0.92}             \\ \hline
Final-871  & \multicolumn{1}{l|}{0.62} & 115.0                    & \multicolumn{1}{l|}{0.62}          & 26.12                     & \multicolumn{1}{l|}{0.62} & 481.22                    & \multicolumn{1}{l|}{0.62} & 76.65                     & \multicolumn{1}{l|}{0.62} & 8.20                      & \multicolumn{1}{l|}{\textbf{0.62}} & \textbf{7.75}             \\ \hline
Final-961  & \multicolumn{1}{l|}{0.94} & 44.30                     & \multicolumn{1}{l|}{0.94}          & 12.82                     & \multicolumn{1}{l|}{0.94} & 710.55                    & \multicolumn{1}{l|}{0.94} & 348.11                    & \multicolumn{1}{l|}{0.94} & 3.32                      & \multicolumn{1}{l|}{\textbf{0.94}} & \textbf{2.85}             \\ \hline
Final-1936 & \multicolumn{1}{l|}{0.89} & 106.0                    & \multicolumn{1}{l|}{0.89}          & 17.03                     & \multicolumn{1}{l|}{0.89} & 789.57                    & \multicolumn{1}{l|}{0.89} & 595.28                    & \multicolumn{1}{l|}{0.89} & 7.03                      & \multicolumn{1}{l|}{\textbf{0.89}} & \textbf{6.67}             \\ \hline
Final-3068 & \multicolumn{1}{l|}{1.09} & 31.60                     & \multicolumn{1}{l|}{1.03}          & 5.49                      & \multicolumn{1}{l|}{1.13} & 26203                     & \multicolumn{1}{l|}{0.98} & 72.27                     & \multicolumn{1}{l|}{1.01} & 2.82                      & \multicolumn{1}{l|}{\textbf{1.01}} & \textbf{2.40}             \\ \hline
Final-4585 & \multicolumn{1}{l|}{0.57} & 417.0                    & \multicolumn{1}{l|}{\textbf{0.56}} & 66.13                     & \multicolumn{1}{l|}{0.56} & 23221                     & \multicolumn{1}{l|}{0.56} & 373.30                    & \multicolumn{1}{l|}{0.57} & 22.22                     & \multicolumn{1}{l|}{0.57}          & \textbf{15.60}                 \\ \hline
\end{tabular}
\caption{The results of the Final dataset.}
\label{tab:final}
\end{table}

%% file: table/PBA.tex
\begin{table}[!t]
\centering
\begin{tabular}{|c|cc|cc|}
 \hline
              & \multicolumn{2}{c|}{PBA(FP32)}    & \multicolumn{2}{c|}{MegBA-1-m(FP32)}               \\ \hline
              & \multicolumn{1}{c|}{MSE}   & Time & \multicolumn{1}{c|}{MSE}           & Time          \\ \hline
Trafalgar-257 & \multicolumn{1}{c|}{0.85}  & 0.57 & \multicolumn{1}{c|}{\textbf{0.44}} & \textbf{0.11} \\ \hline
Ladybug-1695  & \multicolumn{1}{c|}{34.13} & 0.37 & \multicolumn{1}{c|}{\textbf{1.33}} & \textbf{0.32} \\ \hline
Dubrovnik-287 & \multicolumn{1}{c|}{0.84}  & 1.22 & \multicolumn{1}{c|}{\textbf{0.38}} & \textbf{0.10} \\ \hline
Venice-89     & \multicolumn{1}{c|}{0.86}  & 1.32 & \multicolumn{1}{c|}{\textbf{0.51}} & \textbf{0.28} \\ \hline
Final-4585    & \multicolumn{1}{c|}{1.87}  & 4.96 & \multicolumn{1}{c|}{\textbf{0.57}} & \textbf{3.13} \\ \hline
\end{tabular}
\caption{We evaluate PBA with the BAL dataset. To make a fair comparison, we compare PBA against the FP32 MegBA. Both PBA and FP32 MegBA suffer from numerical instability issues in some datasets. In addition, PBA uses texture memory, it failed to solve large-scale BA problems due to memory limitations. We chose the largest datasets that can be solved successfully by PBA to evaluate its performance. The result shows that MegBA outperforms PBA by a large margin, with maximum 12.5 times speed up while achieving a lower MSE.}
\label{tab:PBA}
\end{table}

%% file: table/1dsfm.tex
\begin{table}[!t]
\centering
\begin{tabular}{|l|l|l|}
\hline
Dataset                 & \#Points & \#Observations \\ \hline
Alamo-577               & 140080   & 816891         \\ \hline
Ellis Island-233        & 9210     & 20500          \\ \hline
Gendarmenmarkt-704      & 76964    & 268747         \\ \hline
Madrid Metropolis-347   & 44479    & 195660         \\ \hline
Montreal Notre Dame-459 & 151876   & 811757         \\ \hline
Notre Dame-548          & 224153   & 1172145        \\ \hline
NYC Library-334         & 54757    & 211614         \\ \hline
Piazza del Popolo-336   & 29731    & 150161         \\ \hline
Piccadilly-2292         & 184475   & 798085         \\ \hline
Roman Forum-1067        & 223844   & 1031760        \\ \hline
Tower of London-484     & 126648   & 596690         \\ \hline
Trafalgar-5052          & 327920   & 1266102        \\ \hline
Union Square-816        & 26430    & 90668          \\ \hline
Vienna Cathedral-846    & 154394   & 495940         \\ \hline
Yorkminster-429         & 100426   & 376980         \\ \hline
\end{tabular}
\caption{The statistics of the datasets in 1DSfM.}
\label{tab:1dsfm_statistics}
\end{table}

\begin{table}[!t]
\centering
\begin{tabular}{|l|ll|ll|ll|ll|ll|}
\hline
             & \multicolumn{2}{c|}{Ceres-16}                                    & \multicolumn{2}{c|}{DeepLM}                                      & \multicolumn{2}{c|}{g2o-16}                                      & \multicolumn{2}{c|}{Rootba-16}                                   & \multicolumn{2}{c|}{MegBA-1-m}                                   \\ \hline
             & \multicolumn{1}{c|}{MSE}             & \multicolumn{1}{c|}{Time} & \multicolumn{1}{c|}{MSE}             & \multicolumn{1}{c|}{Time} & \multicolumn{1}{c|}{MSE}             & \multicolumn{1}{c|}{Time} & \multicolumn{1}{c|}{MSE}             & \multicolumn{1}{c|}{Time} & \multicolumn{1}{c|}{MSE}             & \multicolumn{1}{c|}{Time} \\ \hline
Alamo        & \multicolumn{1}{l|}{4.8E+2}          & 219.0                     & \multicolumn{1}{l|}{2.3E+2}          & 40.4                      & \multicolumn{1}{l|}{7.6E+5}          & 48.9                      & \multicolumn{1}{l|}{2.3E+2}          & 31.1                      & \multicolumn{1}{l|}{\textbf{2.2E+2}} & \textbf{18.9}             \\ \hline
Ellis Island & \multicolumn{1}{l|}{2.2E+3}          & 6.3                       & \multicolumn{1}{l|}{2.3E+3}          & 9.9                       & \multicolumn{1}{l|}{4.6E+5}          & 2.7                       & \multicolumn{1}{l|}{\textbf{1.6E+3}} & 1.8                       & \multicolumn{1}{l|}{2.8E+3}          & \textbf{0.5}              \\ \hline
Gen.markt    & \multicolumn{1}{l|}{2.5E+1}          & 63.0                      & \multicolumn{1}{l|}{6.2E+1}          & 31.2                      & \multicolumn{1}{l|}{4.4E+4}          & 46.0                      & \multicolumn{1}{l|}{1.5E+2}          & 20.8                      & \multicolumn{1}{l|}{\textbf{1.6E+1}} & \textbf{2.4}              \\ \hline
M.Metropolis & \multicolumn{1}{l|}{8.9E+3}          & 45.7                      & \multicolumn{1}{l|}{8.4E+3}          & 21.2                      & \multicolumn{1}{l|}{2.8E+8}          & 51.4                      & \multicolumn{1}{l|}{\textbf{8.0E+3}} & 15.8                      & \multicolumn{1}{l|}{1.1E+4}          & \textbf{1.1}              \\ \hline
M.N.Dame     & \multicolumn{1}{l|}{8.4E+2}          & 206.0                     & \multicolumn{1}{l|}{\textbf{7.1E+2}} & 73.4                      & \multicolumn{1}{l|}{5.0E+5}          & 89.1                      & \multicolumn{1}{l|}{7.3E+2}          & 46.4                      & \multicolumn{1}{l|}{7.4E+2}          & \textbf{10.6}             \\ \hline
Notre Dame   & \multicolumn{1}{l|}{\textbf{1.5E+4}} & 282.0                     & \multicolumn{1}{l|}{1.6E+4}          & 92.3                      & \multicolumn{1}{l|}{4.2E+8}          & 68.6                      & \multicolumn{1}{l|}{2.3E+4}          & 48.3                      & \multicolumn{1}{l|}{2.0E+4}          & \textbf{10.8}             \\ \hline
NYC Library  & \multicolumn{1}{l|}{4.8E+0}          & 62.7                      & \multicolumn{1}{l|}{1.2E+1}          & 37.4                      & \multicolumn{1}{l|}{\textbf{2.4E+0}} & 23.2                      & \multicolumn{1}{l|}{3.0E+1}          & 17.7                      & \multicolumn{1}{l|}{4.1E+0}          & \textbf{1.4}              \\ \hline
P.del Popolo & \multicolumn{1}{l|}{7.5E+1}          & 26.5                      & \multicolumn{1}{l|}{8.4E+1}          & 4.6                       & \multicolumn{1}{l|}{5.3E+6}          & 1.4                       & \multicolumn{1}{l|}{\textbf{3.2E+1}} & 19.3                      & \multicolumn{1}{l|}{3.7E+1}          & \textbf{1.4}              \\ \hline
Piccadilly   & \multicolumn{1}{l|}{4.8E+2}          & 213.0                     & \multicolumn{1}{l|}{6.3E+2}          & 68.3                      & \multicolumn{1}{l|}{3.9E+6}          & 21.6                      & \multicolumn{1}{l|}{3.7E+2}          & 73.5                      & \multicolumn{1}{l|}{\textbf{3.7E+2}} & \textbf{14.3}             \\ \hline
R.Forum      & \multicolumn{1}{l|}{6.5E+1}          & 104.0                     & \multicolumn{1}{l|}{5.6E+1}          & 88.5                      & \multicolumn{1}{l|}{1.1E+5}          & 201.6                     & \multicolumn{1}{l|}{5.8E+1}          & 105.8                     & \multicolumn{1}{l|}{\textbf{5.5E+1}} & \textbf{6.0}              \\ \hline
T.of London  & \multicolumn{1}{l|}{1.4E+3}          & 81.9                      & \multicolumn{1}{l|}{1.8E+3}          & 53.5                      & \multicolumn{1}{l|}{6.6E+8}          & 7.1                       & \multicolumn{1}{l|}{1.9E+3}          & 45.9                      & \multicolumn{1}{l|}{\textbf{1.3E+3}} & \textbf{14.7}             \\ \hline
Trafalgar    & \multicolumn{1}{l|}{1.7E+2}          & 358.0                     & \multicolumn{1}{l|}{1.6E+2}          & 112.1                     & \multicolumn{1}{l|}{4.8E+5}          & 34.8                      & \multicolumn{1}{l|}{1.7E+2}          & 211.3                     & \multicolumn{1}{l|}{\textbf{1.5E+2}} & \textbf{15.0}             \\ \hline
Union Square & \multicolumn{1}{l|}{5.4E+2}          & 24.2                      & \multicolumn{1}{l|}{3.9E+2}          & 5.8                       & \multicolumn{1}{l|}{2.9E+6}          & 23.1                      & \multicolumn{1}{l|}{\textbf{2.3E+2}} & 8.9                       & \multicolumn{1}{l|}{3.4E+2}          & \textbf{0.8}              \\ \hline
V.Cathedral  & \multicolumn{1}{l|}{5.7E+2}          & 139.0                     & \multicolumn{1}{l|}{5.2E+2}          & 49.2                      & \multicolumn{1}{l|}{1.1E+5}          & 51.6                      & \multicolumn{1}{l|}{4.9E+2}          & 13.1                      & \multicolumn{1}{l|}{\textbf{4.9E+2}} & \textbf{12.2}             \\ \hline
Yorkminster  & \multicolumn{1}{l|}{1.2E+1}          & 115.0                     & \multicolumn{1}{l|}{9.2E+0}          & 18.9                      & \multicolumn{1}{l|}{\textbf{4.2E+0}} & 25.9                      & \multicolumn{1}{l|}{1.0E+2}          & 29.8                      & \multicolumn{1}{l|}{9.0E+0}          & \textbf{4.4}              \\ \hline
\end{tabular}
\caption{Results of the 1DSfM dataset.}
\label{tab:1dsfm}
\end{table}

%% file: table/1dsfm_mem.tex
\begin{table}[!t]
\centering
\begin{tabular}{|l|l|l|l|l|l|l|}
\hline
             & {ceres16} & {DeepLM} & {g2o} & {RootBA} & {MegBA-1-a} & {MegBA-1-m} \\ \hline
Alamo        & 2.16                         & 1.86                        & 1.96                     & 4.39                        & 2.88                           & \textbf{1.85}                  \\ \hline
Ellis Island & 1.46                         & 1.53                        & 1.15                     & 1.06                        & 0.67                           & \textbf{0.65}                  \\ \hline
Gen.markt    & 1.72                         & 1.83                        & 1.42                     & 1.13                        & 1.36                           & \textbf{1.07}                  \\ \hline
M.Metropolis & 1.62                         & 1.73                        & 1.33                     & 1.14                        & 1.15                           & \textbf{0.95}                  \\ \hline
M.N.Dame     & 2.16                         & 2.40                        & 1.92                     & 3.85                        & 2.87                           & \textbf{1.85}                  \\ \hline
Notre Dame   & 2.50                         & \textbf{2.39}               & 2.29                     & 5.54                        & 3.88                           & 2.47                           \\ \hline
NYC Library  & 1.64                         & 1.76                        & 1.35                     & 1.13                        & 1.20                           & \textbf{0.98}                  \\ \hline
P.del Popolo & 1.57                         & 1.67                        & 1.29                     & 1.31                        & 1.02                           & \textbf{0.89}                  \\ \hline
Piccadilly   & 2.20                         & 2.47                        & 2.18                     & 3.30                        & 2.85                           & \textbf{1.85}                  \\ \hline
R.Forum      & 2.38                         & 2.56                        & 2.17                     & 3.40                        & 3.49                           & \textbf{2.17}                  \\ \hline
T.of London  & 1.99                         & 2.17                        & 1.71                     & 1.80                        & 2.28                           & \textbf{1.55}                  \\ \hline
Trafalgar    & 2.72                         & 2.99                        & 3.15                     & 4.97                        & 4.15                           & \textbf{2.51}                  \\ \hline
Union Square & 1.53                         & 1.58                        & 1.24                     & 1.09                        & 0.86                           & \textbf{0.80}                  \\ \hline
V.Cathedral  & 1.96                         & 2.10                        & 1.67                     & \textbf{1.36}               & 2.00                           & 1.40                           \\ \hline
Yorkminster  & 1.81                         & 2.02                        & 1.52                     & \textbf{1.19}               & 1.67                           & 1.23                           \\ \hline
\end{tabular}
\caption{The memory overhead of the 1DSfM dataset. The unit is GB.}
\label{tab:1dsfmMem}
\end{table}

%% file: figures.tex
\section{Reconstruction Plots}
Finally, we show the visualisation results of \sys. Specifically, we show the landmark point clouds (i.e. black dots) with cameras (i.e. red frames) for each dataset. The point clouds are rendered by COLMAP~\cite{schoenberger2016sfm}\cite{schoenberger2016mvs}.

\begin{figure}[ht!]
\includegraphics[width=8cm]{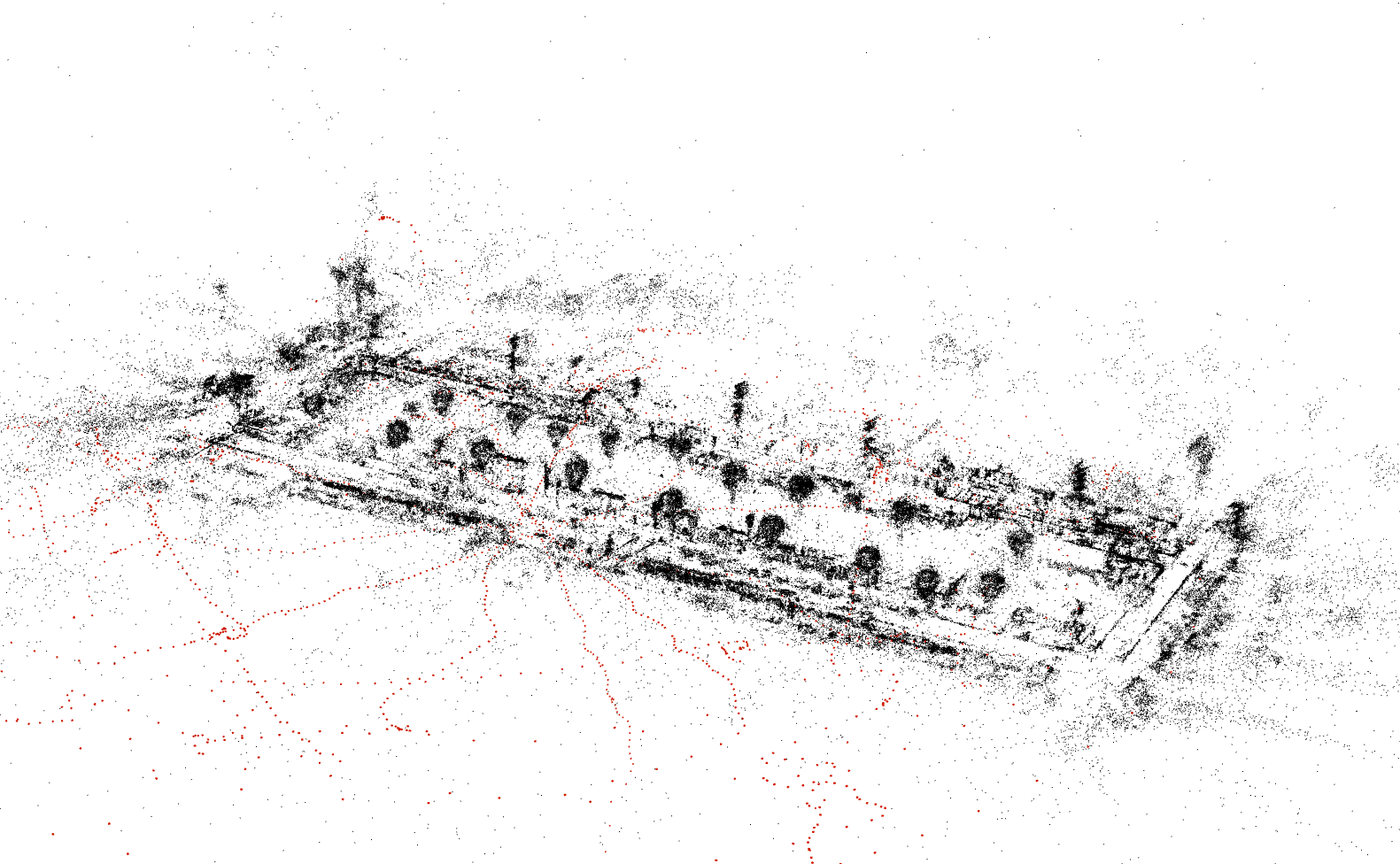}
\centering
\captionsetup{labelformat=empty}
\caption{Ladybug}
\end{figure}

\begin{figure}[ht!]
\includegraphics[width=8cm]{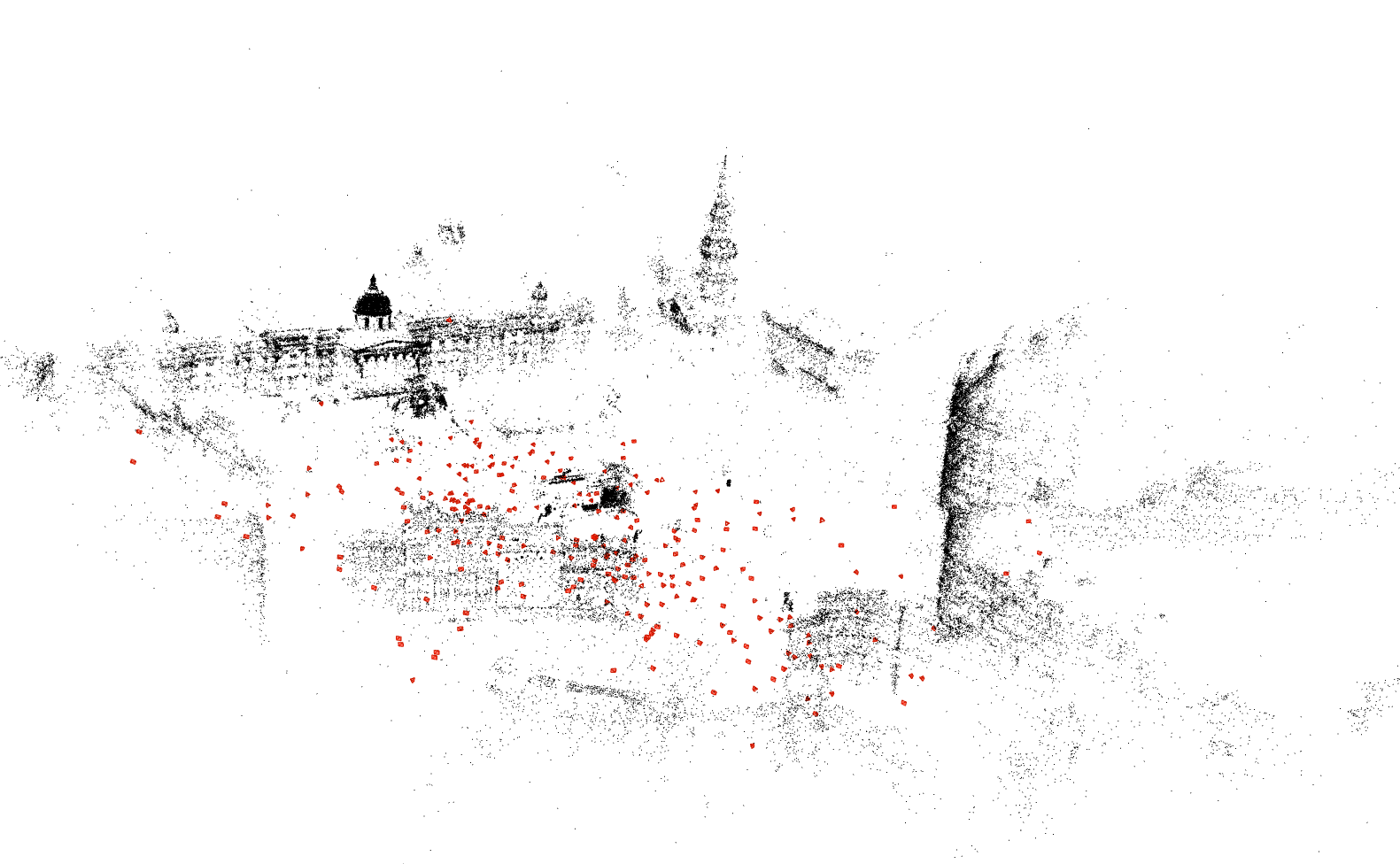}
\centering
\captionsetup{labelformat=empty}
\caption{Trafalgar Square}
\end{figure}

\begin{figure}[ht!]
\includegraphics[width=8cm]{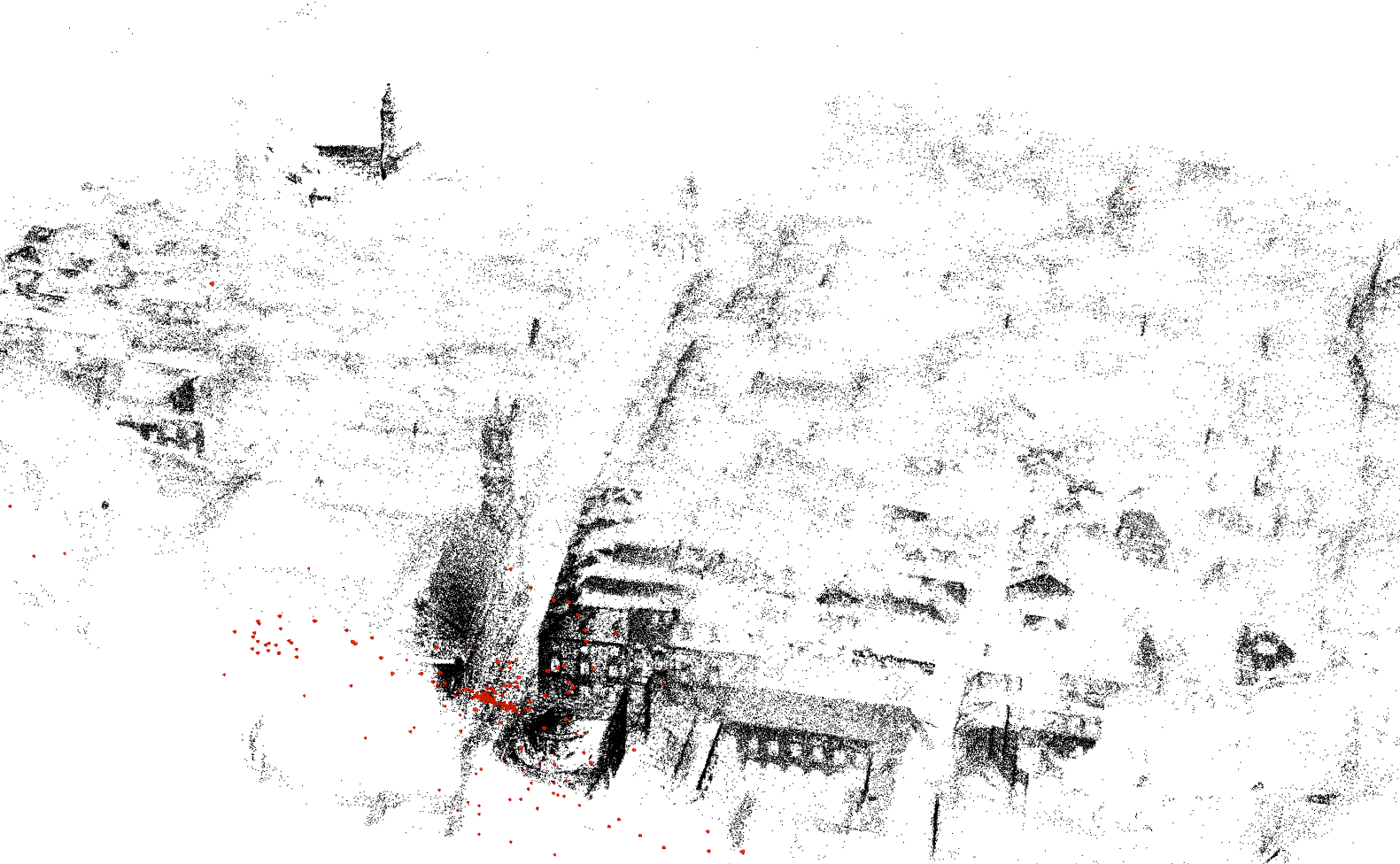}
\centering
\captionsetup{labelformat=empty}
\caption{Dubrovnik}
\end{figure}

\begin{figure}[ht!]
\includegraphics[width=8cm]{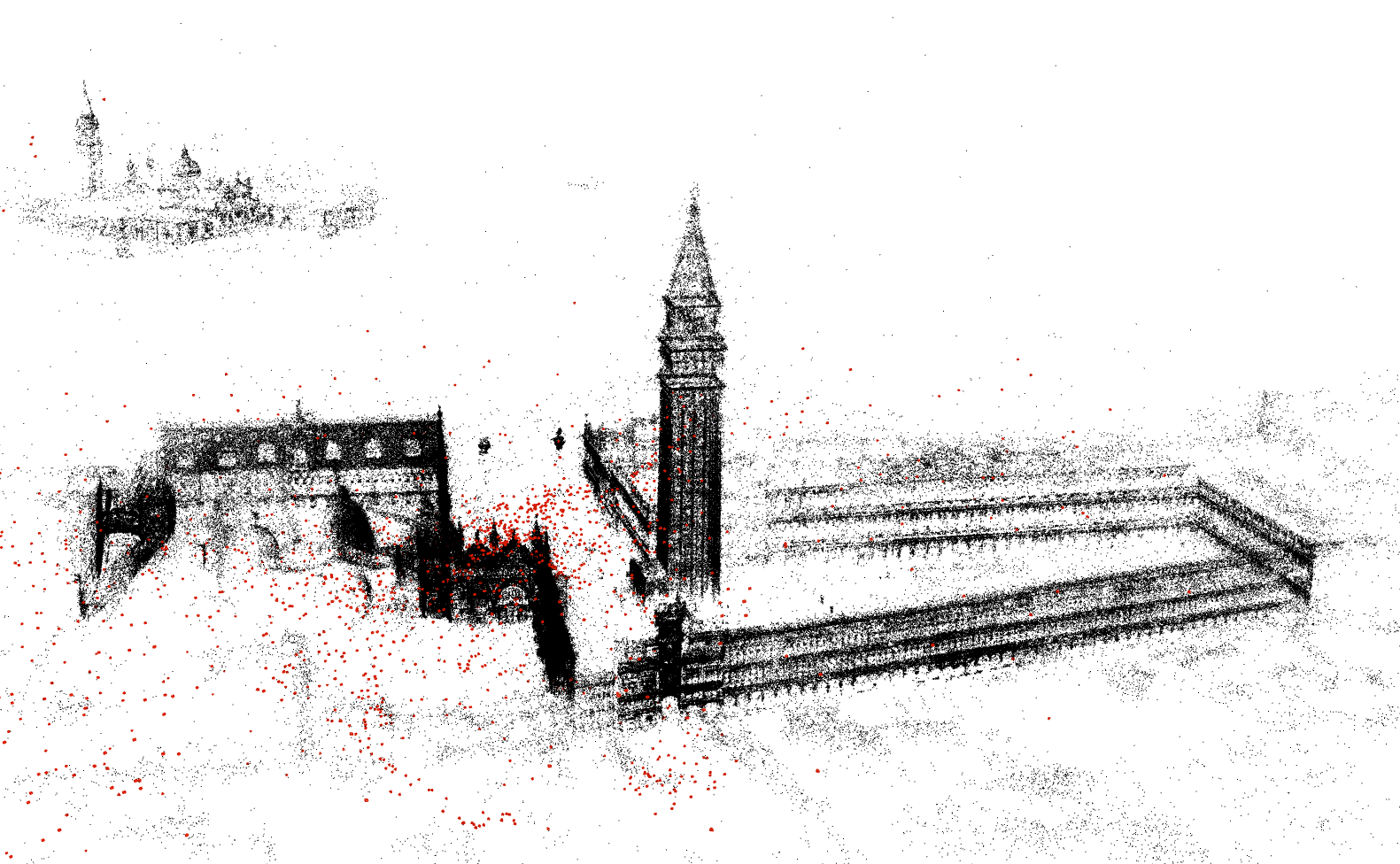}
\centering
\captionsetup{labelformat=empty}
\caption{Venice}
\end{figure}

\begin{figure}[ht!]
\includegraphics[width=8cm]{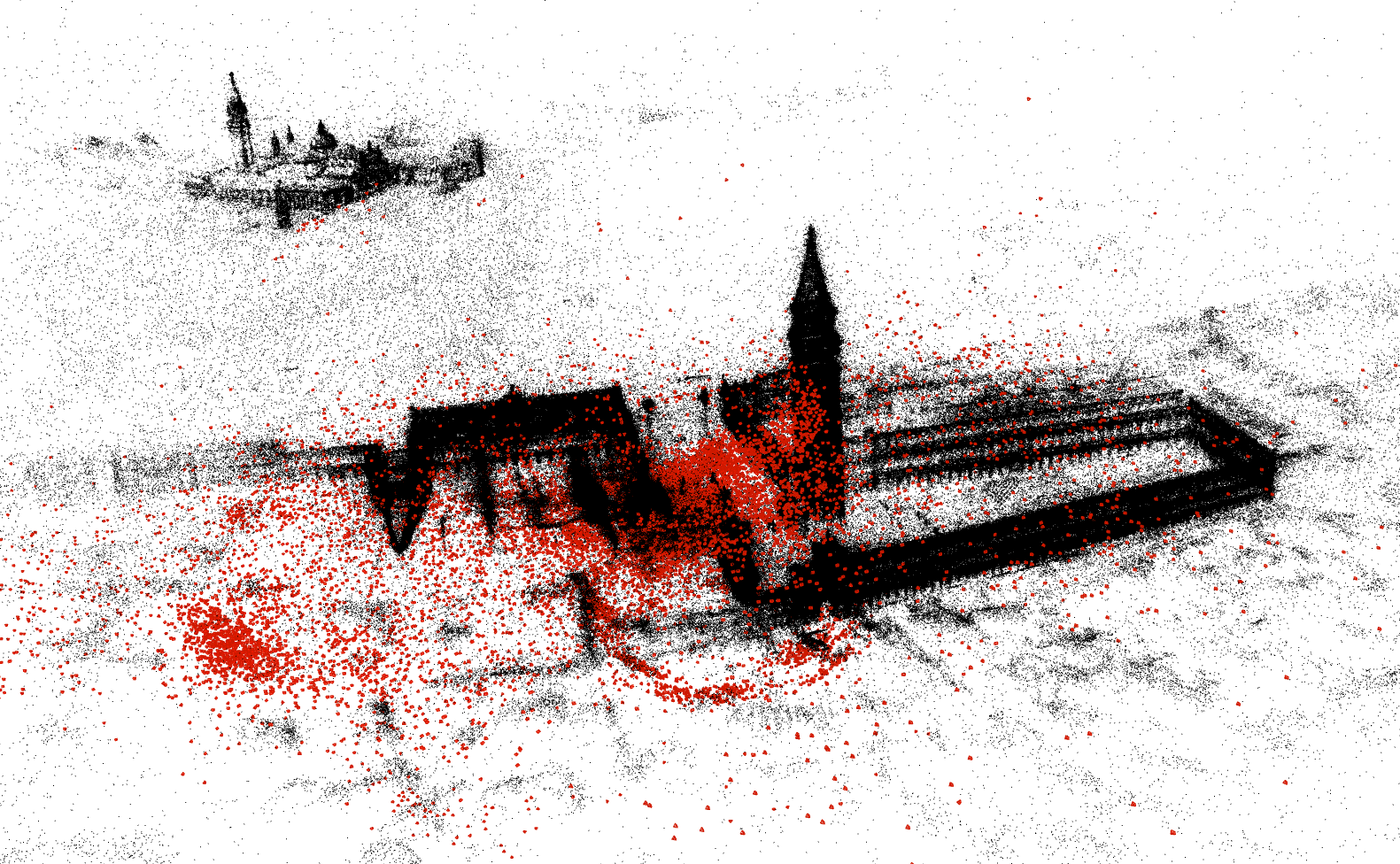}
\centering
\captionsetup{labelformat=empty}
\caption{Final}
\end{figure}